\pdfoutput=1
\documentclass[letterpaper]{article} % DO NOT CHANGE THIS
\usepackage{aaai23}  % DO NOT CHANGE THIS
\usepackage{times}  % DO NOT CHANGE THIS
\usepackage{helvet}  % DO NOT CHANGE THIS
\usepackage{courier}  % DO NOT CHANGE THIS
\usepackage[hyphens]{url}  % DO NOT CHANGE THIS
\usepackage{graphicx} % DO NOT CHANGE THIS
\usepackage{natbib}  % DO NOT CHANGE THIS AND DO NOT ADD ANY OPTIONS TO IT
\usepackage{caption} % DO NOT CHANGE THIS AND DO NOT ADD ANY OPTIONS TO IT
\usepackage{multirow}
\usepackage{booktabs}
\usepackage{bigstrut}
\usepackage{hyperref}
\usepackage{algorithm}
\usepackage{algorithmic}

\usepackage{newfloat}
\usepackage{listings}

\usepackage{amssymb}
\usepackage{amsmath}

\DeclareCaptionStyle{ruled}{labelfont=normalfont,labelsep=colon,strut=off} % DO NOT CHANGE THIS
\floatstyle{ruled}
\newfloat{listing}{tb}{lst}{}
\floatname{listing}{Listing}
\usepackage{color}

\title{One is All: Bridging the Gap Between Neural Radiance Fields Architectures with Progressive Volume Distillation}

\author{
    Shuangkang Fang\textsuperscript{\rm 1}\thanks{Work done during an internship at Megvii.},
    Weixin Xu\textsuperscript{\rm 2},
    Heng Wang\textsuperscript{\rm 2},
    Yi Yang\textsuperscript{\rm 2},
    Yufeng Wang\textsuperscript{\rm 1}\thanks{Corresponding authors.},
    Shuchang Zhou\textsuperscript{\rm 2†}
}
\affiliations{
    \textsuperscript{\rm 1} Beihang University,    \textsuperscript{\rm 2} Megvii Research\\
    \{skfang, wyfeng\}@buaa.edu.cn, \{xuweixin02, wangheng, yangyi, zsc\}@megvii.com
}

\usepackage{bibentry}
\begin{document}

\maketitle

\begin{abstract}
Neural Radiance Fields (NeRF) methods have proved effective as compact, high-quality and versatile representations for 3D scenes, and enable downstream tasks such as editing, retrieval, navigation, etc.  
Various neural architectures are vying for the core structure of NeRF, including the plain Multi-Layer Perceptron (MLP), sparse tensors, low-rank tensors, hashtables and their compositions. 
Each of these representations has its particular set of trade-offs. For example, the hashtable-based representations admit faster training and rendering but their lack of clear geometric meaning hampers downstream tasks like spatial-relation-aware editing. 
In this paper, we propose Progressive Volume Distillation (PVD), a systematic distillation method that allows any-to-any conversions between different architectures, including MLP, sparse or low-rank tensors, hashtables and their compositions. PVD consequently empowers downstream applications to optimally adapt the neural representations for the task at hand in a post hoc fashion. 
The conversions are fast, as distillation is progressively performed on different levels of volume representations, from shallower to deeper. We also employ special treatment of density to deal with its specific numerical instability problem.
Empirical evidence is presented to validate our method on the NeRF-Synthetic, LLFF and TanksAndTemples datasets. For example, with PVD, an MLP-based NeRF model can be distilled from a hashtable-based Instant-NGP model at a 10$\times$$\sim$20$\times$ faster speed than being trained the original NeRF from scratch, while achieving a superior level of synthesis quality. Code is available at \textit{https://github.com/megvii-research/AAAI2023-PVD}.
%measured in PSNR.

\begin{figure}[ht]
\centering
\includegraphics[width=0.4\textwidth]{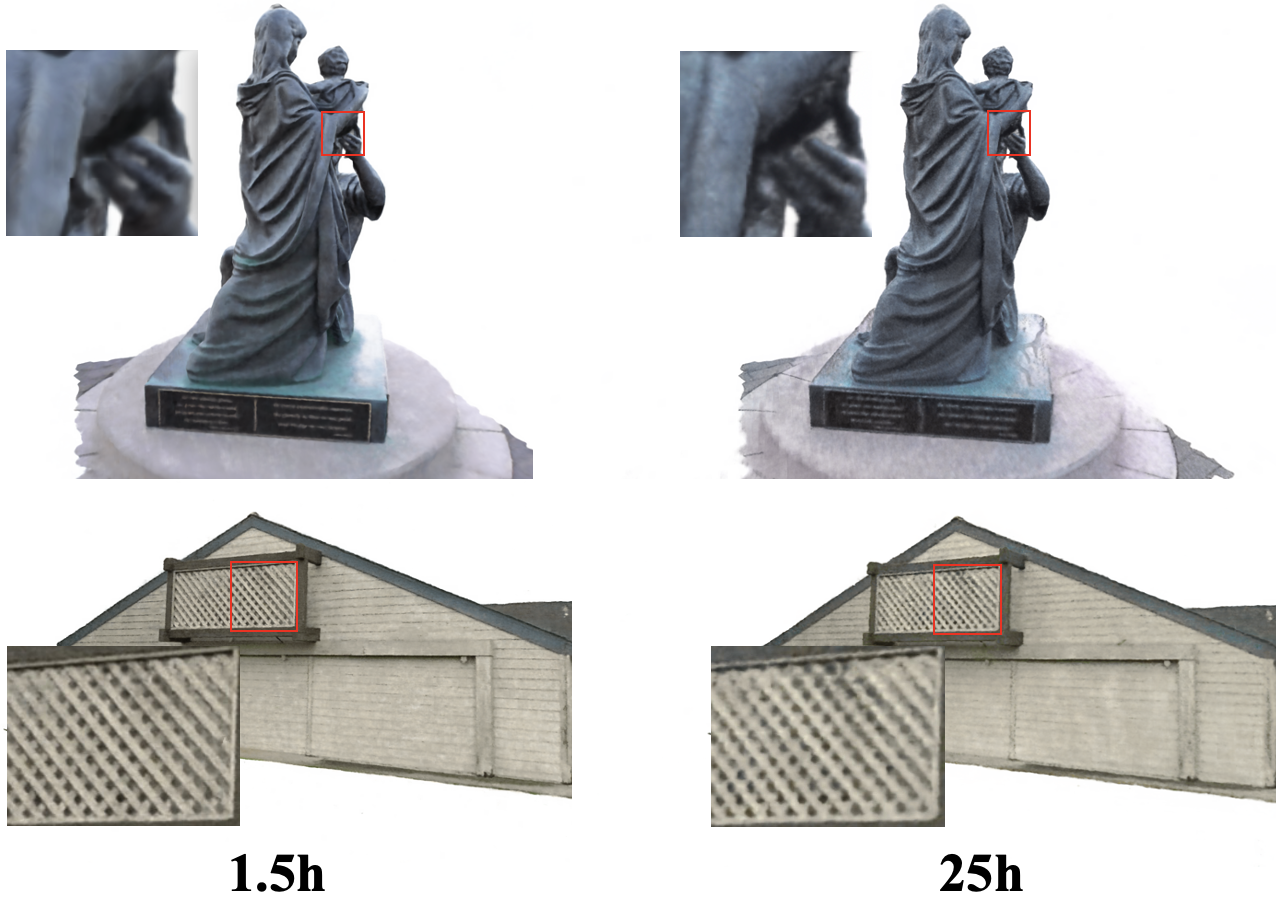} % Reduce the figure size so that it is slightly narrower than the column.
\caption{Comparison of two models trained in the Family and Barn scene from TanksAndTemples dataset. The left is the results of a NeRF model distilled by PVD from an INGP teacher within 1.5 hours. The right is the results of NeRF trained from scratch using 25 hours. PVD improves synthesis quality and reduces training time.}
\label{fig-tank}
\end{figure}

\begin{figure*}[t]
\centering
\includegraphics[width=0.95\textwidth]{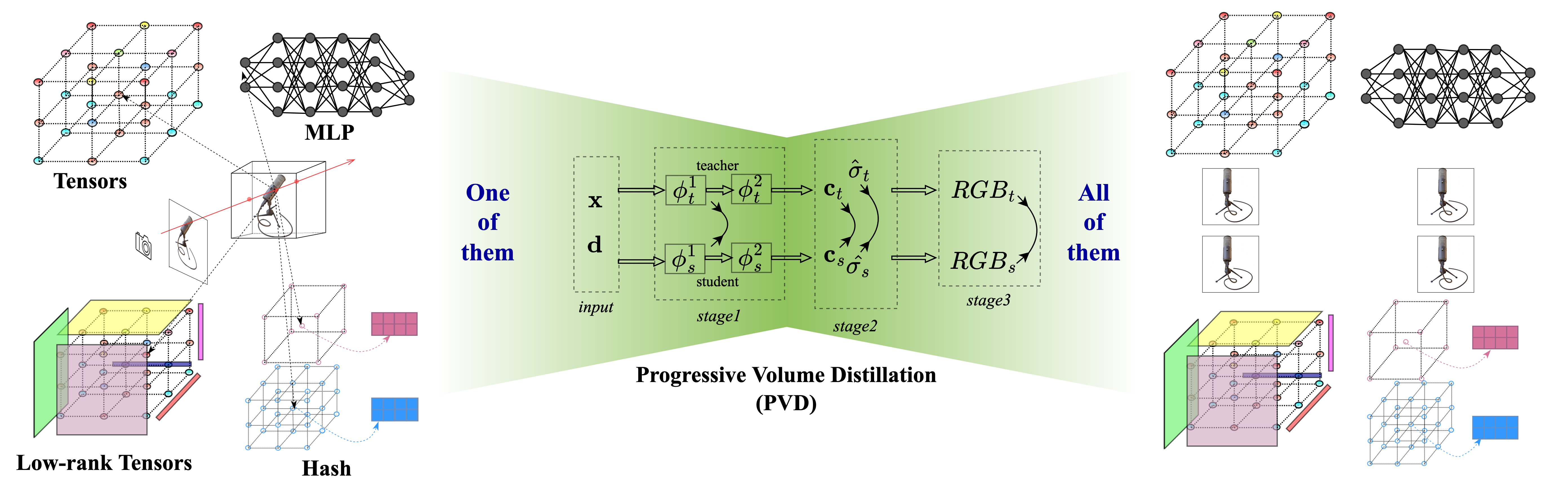} % Reduce the figure size so that it is slightly narrower than the column.
\caption{ With PVD, given one trained NeRF model, different NeRF achitecutres, like sparse tensors, MLP, low-rank tensors and hash tables can be obtained quickly through distillation. The loss in intermediate volume representations (shown as double arrow symbol) like output of $\phi_{*}^1$, color and density are used alongside the final rendered RGB volume to accelerate distillation.}
\label{fig-flow}
\end{figure*}

\end{abstract}

\section{Introduction}
Novel view synthesis (NVS) generates photo realistic 2D images for unknown view-ports of a 3D scene \cite{zhou2018stereo,chan2021pi,sitzmann2019scene}, and has wide applications in rendering, localization, and robot arm manipulations \cite{adamkiewicz2022vision,moreau2022lens,peng2021megloc}, especially with the neural modeling capabilities offered by the recently developed Neural Radiance Fields (NeRF).
%Novel view synthesis(NVS) is a key technology to connect the 2D and 3D scenes. The goal of NVS is to obtain a photo realistic 2D image from a 3D scene when specifying a novel view that the model has never seen before \cite{zhou2018stereo,chan2021pi,sitzmann2019scene}. 
%Recently, the advent of Neural Radiance Fields (NeRF) \cite{mildenhall2020NeRF} makes NVS enter a new period. NeRF represents a 3D scene as a neural network that maps the coordinates and viewing directions of a spatial point into density and color, and then renders to get the 2D images. 
%
Exploiting the strong generalization capabilities of Multi-Layer Perceptrons (MLPs), NeRF can significantly improve the quality of NVS.
% Using the MLP as implicit scene function to build the radiance fields significantly improves the quality of the rendering. Different from the implicit expression, \citeauthor{fridovich2022plenoxels} propose the Plenoxels to express a scene explicitly, which directly learn the density and spherical harmonic coefficients at each voxel in a sparse grid without using any MLP. %
Several following developments incorporate feature tensors as complementary explicit representations to relieve the MLPs from remembering all details of the scene, resulting in faster training speed and more flexible manipulation of geometric structure. The bloated size of the feature tensors in turn spurs works targeting more compact representations, like TensoRF \cite{chen2022tensorf} that leverages VM (vector-matrix) decomposition and canonical polyadic decomposition (CPD), \citeauthor{fridovich2022plenoxels} that exploits the sparsity of the tensor, and Instant Neural Graphics Primitives (INGP) \cite{muller2022instant} that utilizes multilevel hash tables for effective compression of feature tensors.
% Several following works absorb the advantages of the explicit and implicit representation to construct hybrid methods. TensoRF \cite{chen2022tensorf} leverages a VM (vector-matrix) decomposition to decompose a matrix into the low-rank space. Instant-ngp (INGP) \cite{muller2022instant} adopts multiple hash tables to map a grid sparsely. After the explicitly process, TensoRF and INGP utilize an MLP to generate the density and color.

% sitzmann2019scene. This is an implicit modeling method by optimizing an underlying continuous 
% However, NeRF requires dense sampling of spatial points leading to multiple queries of MLP during training, resulting in a slow running speed.
%Although Plenoxels significantly improves the speed of training, the performance of this method depends on a large grid resolution required an immense memory footprint.
All these schemes have their own advantages and limitations. 
Generally, with implicit representations, it would be easier to perform \textit{texture editing} of a scene (such as color, lighting changes and deformations, etc.), to the extent of artistic stylization and dynamic scene modeling \cite{tang2022CCNeRF, kobayashi2022decomposingEditing,pumarola2021dNeRF,gu2021styleNeRF,zhan2021multimodal}. 
% And the structure of MLP is also easily supported by most deep learning chips or devices \cite{xu2021arch, vanholder2016efficient,zhang2020neurochip}. However, it requires dense sampling of spatial points with multiple queries of the MLP during training, resulting in a slow running speed. 
%
On the other hand, methods with explicit or hybrid representation usually enjoy faster training due to the shallower representations and cope better with geometric-aware editing, like merging and other manipulations of scenes, which is in clear contrast to the case of purely implicit representations.

Due to the diversity of downstream tasks of NVS, there is no single answer as to which representation is the best. The particular choice would depend on the specific application scenarios and the available hardware computation capabilities. In this paper, we tackle the problem from another perspective. Instead of focusing on an ideal alternative representation that embraces the advantages of all variants, we propose a method to achieve arbitrary conversions between known NeRF architectures, including MLPs, sparse tensors, low-rank tensors, hash tables and combinations thereof. Such flexible conversions can obviously bring the following advantages.
% Our purpose in this article is to allow conversion between these different representations. 
%
%Perhaps one may ask why do we need to convert them, and what are the benefits? Based on our exploration, we give the following answers. 
%
Firstly, the study would throw insights into the modeling capabilities and limitations of the already rich and ever-growing constellation of architectures of NeRF.
%Firstly, whether the conversion between different forms of models can be carried out is an unexplored question. 
%
Secondly, the possibility of such conversions would free the designer from the burden of pinning down architectures beforehand, as now they can simply adapt a trained model agilely to other architectures to meet the needs of later discovered application scenarios. %different occasions. 
Last but not least, complementary benefits may be leveraged in cases where teacher and student are of different attributes. For example, when a teacher model with hash table is used to distill a student model of explicit representation, it is now possible to benefit from the faster training speed of the teacher while still producing a student model with clear geometric structures.
%explicit or hybrid models are usually trained quickly, while implicit methods are slow. As such, we can leverage an explicit or hybrid model to guide the training of implicit model, whose convergence may be greatly accelerated. 
%
%In addition, the performance of a model can be effectively improved through the guidance of a different form model with high performance. 
% Therefore researchers require to design various forms of models for different situations constantly, and need to train these models from scratch every time, which is very troublesome and inconvenient. We try to mutually transform implicit, explicit and hybrid representations.

The way we realize conversions between different NeRF architectures is PVD, a progressive volume distillation method that operates on different levels of volume representations, from shallower to deeper, with special treatment of the density volume for better numerical stability. In contrast to previous methods proposed for distillation between models of the same architecture, PVD offers any-to-any conversion between possibly heterogeneous NeRF architectures, by first constructing a unified view of them, and then employing a systematic progressive distillation in multiple stages.
%Some methods \cite{reiser2021kiloNeRF, wang2022r2l} for NVS distillation have been proposed, however, they are fundamentally different from our algorithm. They mainly focus on the distillation between models with the same form, while ours highlights the conversion between different forms. The scope of our research is more generic and practical. In order to achieve the distillation between different formal models, we design the multi-loss to project the intermediate layers of different structures into the same implicit space. 
%Then we construct a block-wise scheme to speed up the distillation process. 
%In addition, we impose a limit on the range of density, which significantly improves the performance of distillation. %
Our contributions are summarized as follows.
% we observed that the variance of density is so large that it is difficult to fit directly, so we constraining its range of values during training, significantly improve the performance of distillation by . 
\begin{itemize}
\item We propose PVD, a distillation framework that allows conversions between different NeRF architectures, including the MLP, sparse tensor, low-rank tensor and hash table architectures. To the best of our knowledge, this is the first systematic attempt at such conversions. An array of any-to-any conversion results is presented in Fig.~\ref{fig-mutual}.% to deal with its specific numerical instability problem.
\item In PVD, we build a block-wise distillation strategy to accelerate the training procedure based on a unified view of different NeRF architectures. We also employ a special treatment of the dynamic density volume range by clipping, which improves the training stability and significantly improves the synthesis quality.
%We realize the distillation with high speed and excellent quality.
%
% \item As concrete examples, we perform distillation from XXX
\item As concrete examples, we find that distillation from hashtable and VM-decomposition structures often either helps boost student model synthesis quality and consumes less time than training from scratch. A particular beneficial case, where a NeRF student model is distilled from an INGP teacher, is presented in Fig.~\ref{fig-tank}.

\end{itemize}

\section{Related Work}

\subsection{Neural Implicit Representations} 
Neural implicit representation methods use MLP to construct a 3D scene from coordinate space, as proposed in NeRF \cite{mildenhall2020NeRF}. The input of the MLP is a 5D coordinate (spatial location [$x, y, z$] and viewing direction [$\theta, \phi$], and the output is the volume density and view-dependent color \cite{mildenhall2019llff, sitzmann2019svn, lombardi2019nv, bi2020deep}. The advantage of implicit modeling is that the representation is conducive to controlling or changing texture-like attributes of the scene. For example, \citeauthor{kobayashi2022decomposingEditing} use the pretrained CLIP model \cite{radford2021clip} to induce editing of NeRF representation of a scene. \citeauthor{pumarola2021dNeRF} successfully apply NeRF to the rendering of dynamic scenes by mapping time $t$ to implicit space through an MLP. \citeauthor{martin2021NeRFinthewild} realize the control of scene lighting by adding appearance embedding. However, MLP-based NeRF requires on-the-fly dense sampling of spatial points, which leads to multiple queries of the MLP during training and inference, resulting in slower running speed.%a tardy running. 
% (changing color, lighting, deformation, etc.)
% Abundant subsequent works have continued NeRF and made an improvement of running speed \cite{barron2021mip} propose mip-NeRF, which represents the scene by conical frustums instead of rays in NeRF, achieving faster rendering and better image synthesis quality. \cite{piala2021termiNeRF} propose to add an additional network to predict the boundary position where the camera rays should stop, so as to reduce the number of queries of MLP to achieve effective acceleration. \cite{zhang2020NeRF++} resolve potential ambiguities issue in original NeRF and improves performance on unbounded 3D scenes. \cite{tancik2022block} present block-NeRF which can learning a large-scale environment well. 
%and transient embedding

\subsection{Neural Explicit Representations and Hybrids}
With the explicit representations, the scene is placed directly on a 3D grid (a huge tensor). Each voxel on the grid stores the information of density and color. \citeauthor{fridovich2022plenoxels} first show that a 3D scene can be represented by an explicit grid, and the spherical harmonic coefficients at each voxel can be used to obtain the density and color at arbitrary spatial point by trilinear interpolation. The training and inference speed of Plenoxels is significantly superior to that of MLP-based NeRF. Recently, motivated by the low-rank tensor approximation algorithm, TensoRF \cite{tang2022CCNeRF} decomposes the explicit tensor into low-rank components, which significantly reduces the model size. \citeauthor{rasmuson2022perf} continue to evolve the explicit expression and regard the optimization of grid as a non-linear least squares optimization problem that can be solved more efficiently by Gauss-Newton method. With explicit representation, it is not as easy to make artistic creations as with implicit representation. Nevertheless, explicit representations facility the geometry editing of the scene, including merging of multiple scenes, inpainting and manipulations of objects at specific positions.
% which can be used to render 2D real images.none-sparse spatial points can be obtained .Since no neural network is used, , as well as the editing of combination and separation between scenes.
%\subsection{hybrid Representations}
%hybrid representations have both explicit and implicit structures. 
There are also attempts exploiting a hybrid of the explicit and implicit representations as NeRF architectures \cite{usvyatsov2022t4dt,garbin2021fastNeRF,muller2022instant,chen2022tensorf,wu2022diver}.
The explicit part usually stores features related to the scene, while the implicit part is typically an MLP that interpret the features to obtain densities and colors. Differences between hybrid representations are mainly exhibited in the explicit part. \citeauthor{liu2020nsvf} use a spare grid to store features, while \citeauthor{yu2021plenoctrees} optimize the 3D grid through an octree. \citeauthor{wizadwongsa2021nex} propose an Implicit-Explicit modeling strategy by storing the coefficient as a learnable parameter to accelerate training procedure. Recently, \citeauthor{muller2022instant} propose the multi-resolution hash encoding (MHE), which maps the given coordinate to feature via a cascade of hash tables at different scales. Like TensoRF \cite{chen2022tensorf}, MHE significantly reduces memory footprint and improve inference speed. However, the compactness of MHE comes at a cost of less straightforward geometric interpretation as there are abundant spatial aliasings caused by the hash mechanism.

%Since the explicit structure still exists in the hybrid representation, it also imposes certain limitations on the artistic creation of the scene. 
%, and then interprets these features into density and color through the implicit structure

\subsection{Knowledge Distillation}
Knowledge distillation commonly refers to training a small model to match the output of a larger model (may be trained beforehand or on-the-fly), which is widely used in model optimization and compression \cite{hinton2015distilling, gou2021knowledge}. Multiple attempts have been made in the field of NVS. \citeauthor{barron2022mip} propose an online distillation method to improve the quality of rendering. \citeauthor{wang2022r2l} distill a NeRF model into a model based on neural light fields. The most related to our work is KiloNeRF \cite{reiser2021kiloNeRF}, which uses a huge pretrained NeRF (teacher) to guide thousands of small NeRF models (students) for speeding up. However, KiloNeRF only performs distillation between the same MLP architecture, and the distilling process is significantly slowed down by the continuous querying of the huge MLP in the teacher model.
%While in our method, we overcome such shortcomings in KiloNeRF by distillation between different representations. 
% We can distill a NeRF model in a short time by using a teacher based, an explicit or hybrid representation.
%  to a small model has obvious speed defects as the training of the huge NeRF. In order to achieve the effect of scene editing and artistic creation, the outputs of, In this article, we mainly focus on the distillation between models with different representations.

\section{Method}
\label{sec:method}
Our method aims to achieve mutual conversions between different architectures of Neural Radiance Fields. Since there is an ever-increasing number of such architectures, we will not attempt to achieve these conversions one by one. Rather, we first formulate typical architectures in a unified form and then design a systematic distillation scheme based on the unified view. The architectures we have derived formula include implicit representations like MLP in NeRF, explicit representations like sparse tensors in Plenoxels, and two hybrid representations: hash tables (in INGP) and low-rank tensors (VM-decomposition in TensoRF). Once formulated, any-to-any conversion between these architectures and their compositions is possible. We will first cover some preliminaries before moving to a detailed description of our method. 
%And the reason we chose two hybrid representations is the enormous gap in their explicit parts, and we expect to realize the conversion between Hash and VM-decomposition.

\subsection{Preliminaries}
\subsubsection{Neural Radiance Fields}
NeRF represents scenes with an implicit function that maps spatial point $\mathbf{x}=(x, y, z)$ and view direction $\mathbf{d}=(\theta, \phi)$ into the density $\sigma$ and color $\mathbf{c}$. Given a ray $\mathbf{r}$ originating at $\mathbf{o}$ with direction $\mathbf{d}$, the RGB value $\hat{\mathbf{C}}(\mathbf{r})$ of the corresponding pixel is estimated by the numerical quadrature of the color $\mathbf{c}_i$ and density $\sigma_i$ of the spatial points $\mathbf{x}_i=\mathbf{o}+t_i\mathbf{d}$ sampled along the ray:
%\cite{mildenhall2020NeRF} 
\begin{equation}  \label{neural_rendering_equation}
    \hat{\mathbf{C}}(\mathbf{r}) = \sum_{i}^{N}T_i(1-\exp(-\sigma_i \delta_i))\mathbf{c}_i
\end{equation}
where $T_i = \exp(- \sum_{j=1}^{i-1} \sigma_i \delta_i)$, and $\delta_i$ is the distance between adjacent samples. 
%=t_{i+1}-t_i

\subsubsection{Tensors and Low-rank Tensors} The Plenoxels directly represents a 3D scene by an explicit grid (tensor) \cite{fridovich2022plenoxels}. Each grid point stores density and spherical harmonic (SH) coefficients. The color $c$ is obtained according to the SH and the view direction $\mathbf{d}$ as follows:
\begin{equation}  \label{eq-sh}
c(\mathbf{d} ; \mathbf{k})=S\left(\sum_{\ell=0}^{\ell_{\max }} \sum_{m=-\ell}^{\ell} k_{\ell}^{m} Y_{\ell}^{m}(\mathbf{d})\right)
\end{equation}
where $S: x \mapsto(1+\exp (-x))^{-1}$, $\mathbf{k}=\left(k_{\ell}^{m}\right)_{\ell: 0 \leq \ell \leq \ell_{\max }}^{m:-\ell \leq m \leq \ell}$,  and $k_{\ell}^{m}$ is a set of coefficients, and $l$ is the degree of the SH function $Y_{\ell}^{m}$.
%and each voxel stores 1 density and 27 spherical harmonic coefficients (degree=2). 
% In our experiment, the size of the grid is [128, 128, 128], Therefore, the size of the whole tensor is [128, 128, 128, 28].
% The information of none-sparse spatial points can be obtained by trilinear interpolation using the voxels on the sparse grid. 
%  corresponding to the RGB components

The performance of explicit sparse tensors depends excessively on the spatial resolution of the grid. In order to reduce the memory footprint caused by the enormous size of the tensor, The VM (Vector-Matrix) \cite{chen2022tensorf} decomposition factorizes the huge tensor $\mathcal{T} \in \mathbb{R}^{I \times J \times K}$ into low-rank matrices $\mathbf{M}$ and vectors $\mathbf{v}$ as follows:
% the low-rank method is proposed. . A huge tensor  can be approximated by  in VM decomposition 
\begin{equation}  \label{eq-vm}
\mathcal{T}=\sum_{r=1}^{R_{1}} \mathbf{v}_{r}^{1} \circ \mathbf{M}_{r}^{2,3}+\sum_{r=1}^{R_{2}} \mathbf{v}_{r}^{2} \circ \mathbf{M}_{r}^{1,3}+\sum_{r=1}^{R_{3}} \mathbf{v}_{r}^{3} \circ \mathbf{M}_{r}^{1,2}
\end{equation}
where $\mathbf{v}_{r}^{1} \in \mathbb{R}^{I}$, $\mathbf{v}_{r}^{2} \in \mathbb{R}^{J}$, $\mathbf{v}_{r}^{3} \in \mathbb{R}^{K}$, $\mathbf{M}_{r}^{2,3} \in \mathbb{R}^{J \times K}$, $\mathbf{M}_{r}^{1,3} \in \mathbb{R}^{I \times K}$, and $\mathbf{M}_{r}^{1,2} \in \mathbb{R}^{I \times J}$. And $\circ$ represents the outer product. Unlike Plenoxels, VM decomposition does not store density and color directly but features that can be decoded by an MLP.

\subsubsection{Multi-resolution Hash Encoding}INGP \cite{muller2022instant} maps a series of grids of different scales to the corresponding feature vectors with fixed size. INGP uses a hash function as in Equation (\ref{eq-hash}) to map a spatial point in the grid to a hash table with different resolution that is adopted to details of different levels of these grids.
%Instead of directly training the voxel on the grid,
\begin{equation}  \label{eq-hash}
h(\mathbf{x})=\left(\bigoplus_{i=1}^{d} x_{i} \pi_{i}\right) \quad \bmod \quad S
\end{equation}
where $\bigoplus$ denotes bit-wise XOR operation. $\pi_{i}$ is an unique large prime number. And $S$ is the hash table size. These hash tables store learnable parameters, which are fed to a shallow MLP to interpret densities and colors. INGP effectively reduces the model size by these hash tables and improves the synthesis quality by introducing multi-resolution.
% In addition, due to the introduction of multi-resolution, the synthesis quality is significantly improved.Since the size of hash tables is much smaller than the size of grids, 
%  Then the output density and color of each point can then be used for neural rendering based on Equation (\ref{neural_rendering_equation}).

\begin{figure*}[t]
\centering
\includegraphics[width=0.85\textwidth]{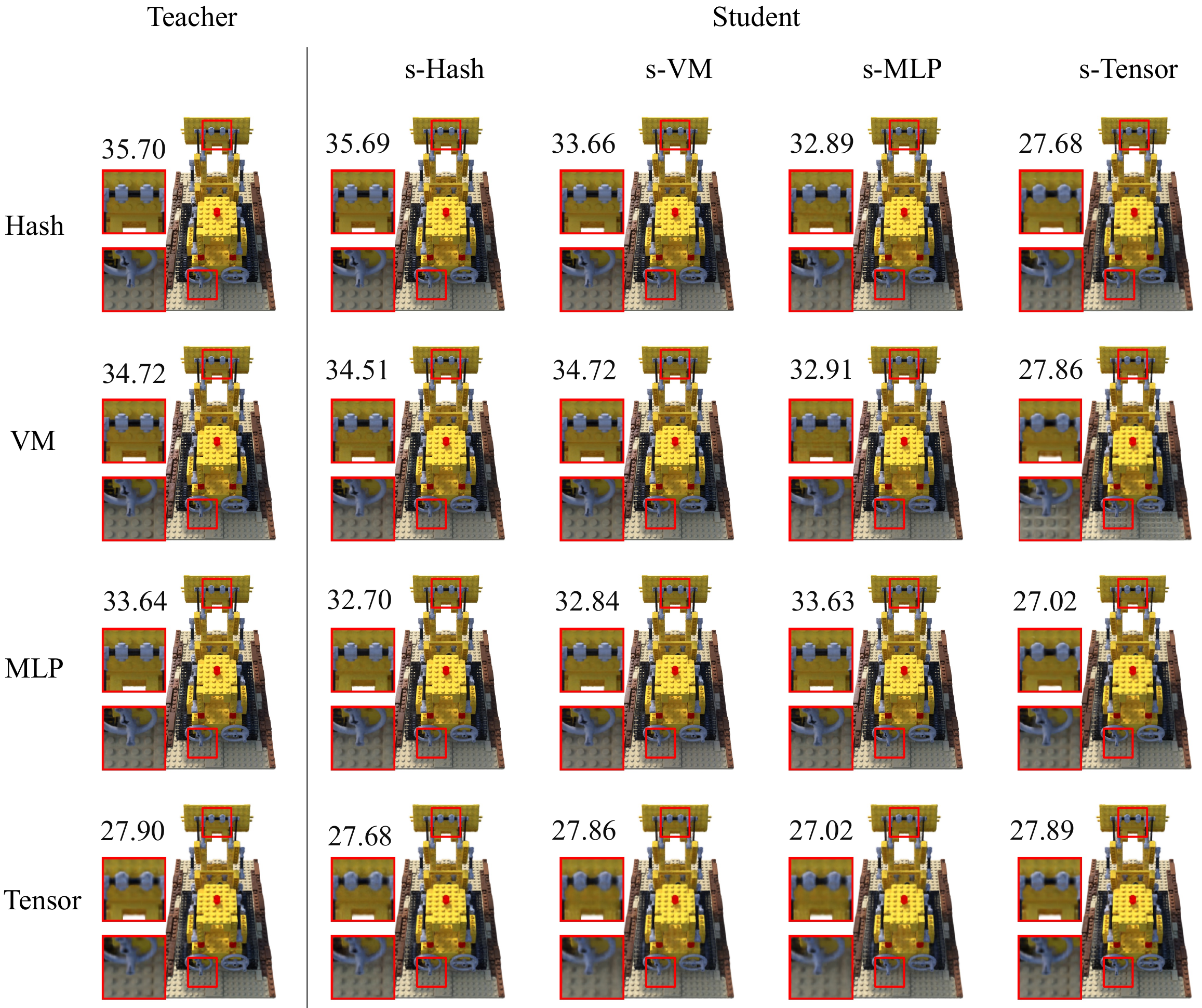} % Reduce the figure size so that it is slightly narrower than the column.
\caption{Quantitative and qualitative results of mutual-conversion between Hash / VM-decomposition / MLP / sparse tensors on the Lego scene from the NeRF-Synthetic dataset. We first train a teacher model for each structure, then use them to distill the student models. The numbers indicate PSNR of the quality of the synthesis. See the supplementary material for more results.}
\label{fig-mutual}
\end{figure*}

\subsection{PVD: Progressive Volume Distillation}
Next we outline the details of PVD. Given a trained model, our task is to distill it into other models, possibly with different architectures. In PVD, we design a volume-aligned loss and build a blockwise distillation strategy to accelerate the training procedure based on a unified view of different NeRF architectures. We also employ a special treatment of the dynamic density volume range by clipping, which improves the training stability and significantly improves the synthesis quality.
% multiple losses to help the distillation procedure. We adopt the block-wise strategy to dynamically use different types of loss so that the model can converge apace without forwarding the whole model at some stages. In addition, we restrict the value range of density to make its fitting more accurate, which is very helpful for high-quality distillation results. 
The illustration of our method is shown in Fig.~\ref{fig-flow}.
% We observed that diverse types of loss play different roles in different training periods. Therefore,
%they just let the final output of student to imitate the output of teacher
\subsubsection{Loss Design}
 In our method, we not only use the RGB, but also use the density, color and an additional intermediate feature to calculate loss between different structures. We observed that the implicit and explicit structures in the hybrid representation are naturally separated and correspond to different learning objectives. Therefore, we consider splitting a model into this similar expression forms so that different parts can be aligned during distillation. Specifically, given a model $\phi_*$, we represent them as a cascade of two modules as follows:
 \begin{equation}
    \phi_*(\mathbf{x}, \mathbf{d}) = \phi_{*}^2(\phi_{*}^1(\mathbf{x}, \mathbf{d}))
 \end{equation}
 
\begin{table}[h]
\centering
\begin{tabular}{ccccc}
\hline
methods & $\phi_{*}^1$ & $\phi_{*}^2$ \\ \hline
NeRF   & first K layers & remaining MLP \\
INGP & hash tables & MLP decoder \\
TensoRF & decomposed tensors & MLP decoder \\
Plenoxels & full & identity function \\ \hline
\end{tabular}
\caption{The division of each architecture under our unified two-level view. Regarding NeRF, K=4 is used by default in this paper.}
\label{table-split}
\end{table}
Here * can be either a teacher or a student. For hybrid representations, we directly regard the explicit part as $\phi_{*}^1$, and the implicit part as $\phi_{*}^2$. While for purely implicit representation, we divide the network into two parts with similar number of layers according to its depth, and denote the former part as $\phi_{*}^1$ and the latter part as $\phi_{*}^2$. As for the purely explicit representation Plenoxels, we still formulate it into two parts by letting $\phi_{*}^2$ be the identity, though it can be transformed without splitting. The specific splitting of the model is shown in Table~\ref{table-split}. Based on the splitting, we design volume-aligned losses as follows:
 \begin{equation}
    \mathcal{L}_{2}^{v} = \left \| \phi_{t}^1(\mathbf{x}, \mathbf{d}) -\phi_{s}^1(\mathbf{x}, \mathbf{d}) \right \| _{2}
\end{equation}
 % While for explicit representation, there is no any 
 % A simple way to do this is to keep the MLP of the student and the teacher same to ensure that the dimension of features are consistent. But we do not want to interfere too much with the structure of the student, so we just add a simple linear layer to a certain position in the student network to make its output dimension consistent with the other one. 
 In essence, the reason for designing this loss is that models in different forms can be mapped to the same space that represents the scene. Our experiments have shown that this volume-aligned loss can accelerate the distillation and improve the quality significantly. Our complete loss function during distillation is as follows:
 % Although \cite{wang2022r2l} and \cite{reiser2021kiloNeRF} have proposed some methods about how to distill a NeRF model, they only use the RGB output of the model to calculate loss while ignoring the effective information of the middle layer.
%of a sampling point and the RGB value finally rendered on the ray to calculate the loss,
\begin{equation}
\mathcal{L}= \omega_{1}\mathcal{L}_{2}^{v} + \omega_{2}\mathcal{L}_{2}^{\sigma}  + \omega_{3}\mathcal{L}_{2}^{c} + \omega_{4}\mathcal{L}_{2}^{rgb} + \omega_{5}\mathcal{L}_{reg}
\end{equation}
where $\mathcal{L}^\sigma,\mathcal{L}^c,\mathcal{L}^{rgb},$ denote the density loss, color loss and RGB loss respectively. $\mathcal{L}_{2}$ is the mean-squared error (MSE). The last item $\mathcal{L}_{r e g}$ represents the regularization term, which depends on the form of the student model. For Plenoxels and VM-decomposition, we add L1 sparsity loss and total variation (TV) regularization loss.
It should be noted that we only perform density, color, RGB and regularization loss on Plenoxels for its explicit representation. Please refer to supplementary materials for more details.
%And we retain the SH coefficients in Plenoxels.  
% the scheme of using spherical harmoni.At this time, the color loss will not be calculated until the spherical harmonic coefficients are changed to color.the structure of Plenoxels based on a pure tensor does not contain any MLP, so
% TODO

\subsubsection{Density Range Constrain}
We found that the loss of density $\sigma$ is hardly directly optimized. And we impute this problem to its specific numerical instability. That is, the density reflects the light transmittance of a point in the space. When $\sigma$ is greater than or less than a certain value, its physical meaning is consistent (i.e., completely transparent or completely opaque). Therefore the value range of $\sigma$ can be too wide for a teacher, but in fact, only one interval of the density values play a key role (a more detailed analysis is in the supplementary material). On the basis of this, we limit the numerical range of $\sigma$ to [$a, b$]. Then the $\mathcal{L}_{2}^{\sigma}$ is calculated as follow:
\begin{equation}
\mathcal{L}_{2}^{\sigma} = \left \| \min (\max (\sigma_{t}, a), b) - \min (\max (\sigma_{s}, a), b) \right \|_2
\end{equation}
According to our experiments, this restricting has an inappreciable impact on the performance of teacher and bring a tremendous benefit to the distillation. We also consider to directly perform the density loss on the $\exp(-\sigma_{i}\delta_{i})$, but we found it is an inefficiency way since the gradient of $\exp$ are easier to saturate, and it requires computing an exponent that increases the amount of calculation when the block-wise is implemented.
%, and a small amount of deviation will cause a large difference in results when using it for subsequent rendering . 
%
% To solve this problem, we consider the actual physical meaning of density and limit its value to a certain range.
% if we directly consistent the densities between student and teacher
% TODO: density range and exp(sigma) in supply

\subsubsection{Block-wise Distillation}
During volume rendering, most of the computation occurs in MLP forwarding for each sampled point and integrating the output over each ray. Such a heavy process slows down the training and distillation significantly.
While in our PVD, thanks to the designed of $\mathcal{L}_{2}^{v}$, we can implement the block-wise strategy to get rid of this problem. Specifically, we only forward stage1 at the beginning of training, and then run stage2 and stage3 in turn as shown in Fig.\ref{fig-flow}. Consequently, the student and the teacher do not need to forward the complete network and render RGB in the early stages of training. In our experiment, the conversion from INGP to NeRF can be completed in tens of minutes, which requires several hours in the past.
%Our method can make full use of the designed multi-loss. thus accelerating the distillation procedure. In addition, if the designer considers sharing the MLP of student and teacher, the training procedure will be faster.in the early stage of distillation, We just align the output of the middle layer of teacher and student.

\begin{figure}[t]
\centering
\includegraphics[width=0.35\textwidth]{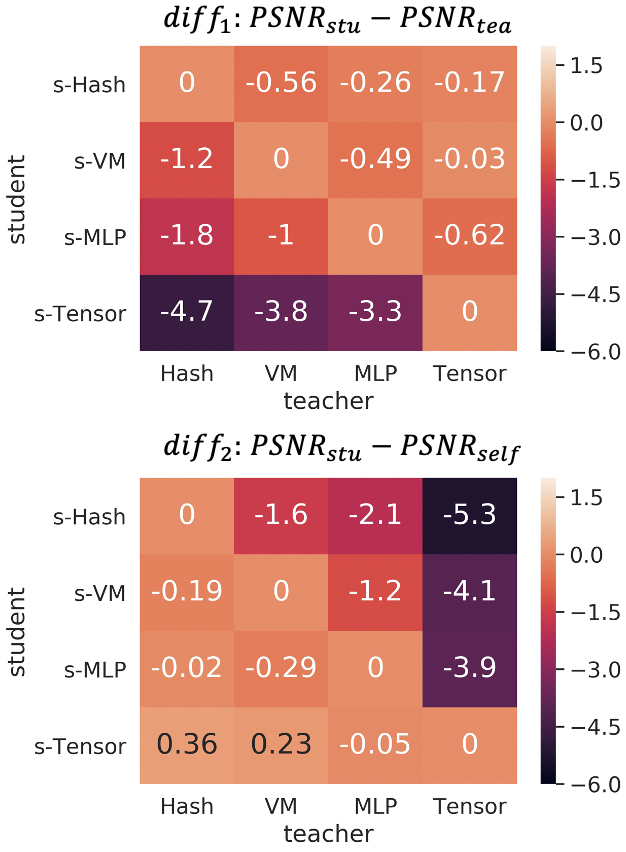} % Reduce the figure size so that it is slightly narrower than the column.
\caption{Gaps in PSNR of mutual-conversion in Synthetic-NeRF dataset. $PSNR_{stu}$ indicates the PSNR of student obtained by distillation. $PSNR_{self}$ represents the PSNR of student obtained by training it from scratch. $PSNR_{tea}$ is the PSNR of the teacher.}
\label{fig-heatmap}
\end{figure}
% TODA: fake data
%. The smaller the value, the smaller the gap between the student and the teacher or itself trained from scratch. Black font indicates that the performance of student exceeds itself trained from scratch
\section{Experiments}
\subsection{Implementation Details}
\subsubsection{Dataset.}Our experiments are mainly carried out on the following three datasets: NeRF-Synthetic dataset  \cite{mildenhall2020NeRF}, forward-facing dataset (LLFF) \cite{mildenhall2019llff} and TanksAndTemple dataset \cite{knapitsch2017tanks}. We only use the above datasets for the training of teacher models. In the distillation stage, we find it sufficient to utilize the teacher to generate fake data as in \textit{pseudo-labeling}, and not touch any of the training data.
\subsubsection{Network Architecture.}
For each structure (Hash / MLP / VM-decomposition / sparse tensors), we keep consistent with their original settings as much as possible. For MLP \cite{lin2020NeRFpytorch}, we also use positional encoding for coordinates and view directions. For sparse tensors \cite{fridovich2022plenoxels}, we use spherical harmonics of degree 2, and the $128\times128\times128$ grid for NeRF-Synthetic dataset and TankAndTemple dataset, $512\times512\times128$ grid for LLFF dataset. For VM-decomposition \cite{chen2022tensorf}, we take 48 components totally. For Hash \cite{muller2022instant}, we set the coarsest resolution, the finest resolution, levels, hash table size and feature dimensions to 16, $2048 \times \text{scene size}$, 14, $2^{19}$, and 2 respectively. 
% TODO
% For convenience, we just use the explicit structure and do not perform a series of up sampling and grid clipping steps in plenoxels.
% Plenoxels will occupy too much memory storage when its grid has large size. For convenience, we only set its grid to 128 for all datasets.

% \usepackage{multirow}
\begin{table*}[t]
\centering
% \usepackage{multirow}
\begin{tabular}{c|cccccccccccc}
\hline
                         & \multicolumn{12}{c}{Teacher}                                                                                                                                                                                                                                                                                                                                                                                                                                                                                                                                                                                                                                                                                             \\ \cline{2-13} 
\multirow{2}{*}{student} & \multicolumn{4}{c|}{PSNR$\uparrow$}                                                                                                                                                                                                                     & \multicolumn{4}{c|}{SSIM$\uparrow $}                                                                                                                                                                                                                     & \multicolumn{4}{c}{$\text{LPIPS}_{Alex}\downarrow$}                                                                                                                                                                                           \\
                         & \begin{tabular}[c]{@{}c@{}}Hash\\ 32.58\end{tabular} & \begin{tabular}[c]{@{}c@{}}VM\\ 31.52\end{tabular} & \begin{tabular}[c]{@{}c@{}}MLP\\ 30.78\end{tabular} & \multicolumn{1}{c|}{\begin{tabular}[c]{@{}c@{}}Tensors\\ 27.49\end{tabular}} & \begin{tabular}[c]{@{}c@{}}Hash\\ 0.960\end{tabular} & \begin{tabular}[c]{@{}c@{}}VM\\ 0.955\end{tabular} & \begin{tabular}[c]{@{}c@{}}MLP\\ 0.946\end{tabular} & \multicolumn{1}{c|}{\begin{tabular}[c]{@{}c@{}}Tensor\\ 0.917\end{tabular}} & \begin{tabular}[c]{@{}c@{}}Hash\\ 0.032\end{tabular} & \begin{tabular}[c]{@{}c@{}}VM\\ 0.040\end{tabular} & \begin{tabular}[c]{@{}c@{}}MLP\\ 0.049\end{tabular} & \begin{tabular}[c]{@{}c@{}}Tensor\\ 0.122\end{tabular} \\ \hline
s-Hash                     & 32.58                                                & 30.96                                              & 30.52                                               & \multicolumn{1}{c|}{27.32}                                                  & 0.960                                                & 0.949                                              & 0.944                                               & \multicolumn{1}{c|}{0.913}                                                  & 0.032                                                & 0.047                                              & 0.053                                               & 0.119                                                  \\
s-VM                       & 31.33                                                & 31.52                                              & 30.29                                               & \multicolumn{1}{c|}{27.46}                                                  & 0.954                                                & 0.955                                              & 0.944                                               & \multicolumn{1}{c|}{0.916}                                                  & 0.042                                                & 0.040                                              & 0.056                                               & 0.121                                                  \\
s-MLP                      & 30.76                                                & 30.49                                              & 30.78                                               & \multicolumn{1}{c|}{26.87}                                                  & 0.946                                                & 0.945                                              & 0.946                                               & \multicolumn{1}{c|}{0.906}                                                  & 0.056                                                & 0.055                                              & 0.049                                               & 0.127                                                  \\
s-Tensors                   & 27.85                                                & 27.72                                              & 27.44                                               & \multicolumn{1}{c|}{27.49}                                                  & 0.921                                                & 0.921                                              & 0.918                                               & \multicolumn{1}{c|}{0.917}                                                  & 0.100                                                & 0.099                                              & 0.098                                               & 0.122                                                  \\ \hline
\end{tabular}
\caption{The qualitative results(PSNR / SSIM / $\text{LPIPS}_{Alex}$) of mutual-conversion between Hash / VM-decomposition / MLP / sparse tensors representations on NeRF-Synthetic dataset. The top number of each column represents the metric of the teacher, and the four numbers below represent the metric of the student obtained by distillation from the teacher. The s- means distillation.}
\label{table-mutual}
\end{table*}

\begin{table*}[t]
\centering
% \usepackage{multirow}
\begin{tabular}{c|cccc|cccc}
\hline
\multirow{2}{*}{method} & \multicolumn{4}{c|}{TanksAndTemple}                                            & \multicolumn{4}{c}{LLFF}                                                       \\ \cline{2-9} 
                        & PSNR           & SSIM           & $\text{LPIPS}_{Alex}$ & $\text{LPIPS}_{VGG}$ & PSNR           & SSIM           & $\text{LPIPS}_{Alex}$ & $\text{LPIPS}_{VGG}$ \\ \hline
Teacher-Hash            & 29.26          & 0.915          & 0.134                 & 0.106                & 26.70          & 0.832          & 0.231                 & 0.130                \\ \hline
TensoRF-VM              & \textbf{28.06} & \textbf{0.909} & \textbf{0.145}        & \textbf{0.155}       & \textbf{26.51} & \textbf{0.832} & \textbf{0.217}        & \textbf{0.135}       \\
Ours: s-VM              & 27.86          & 0.899          & 0.176        & 0.181                & 25.73          & 0.793          & \textbf{0.195}        & 0.269                \\ \hline
NeRF                    & 25.78 & 0.864 & -                     & -                    & \textbf{26.50} & \textbf{0.811} & 0.250                 & -                    \\
Ours: s-MLP             & \textbf{27.50} & \textbf{0.891} & \textbf{0.194}        & 0.190                & 25.77 & 0.784 & \textbf{0.213}        & 0.310                \\ \hline
Plenoxels               & 25.18 & 0.865 & \textbf{0.219}        & 0.261                & \textbf{21.69} & \textbf{0.607} & \textbf{0.527}        & 0.527                \\
Ours: s-Tensors          & \textbf{25.31} & \textbf{0.866} & 0.263                 & \textbf{0.220}       & 21.36 & 0.600 & 0.561                 & \textbf{0.524}       \\ \hline
\end{tabular}
\caption{Comparison of the qualitative results of models (s-VM, s-MLP, s-Tensors) obtained by our distillation method with the models (TensoRF-VM, NeRF, Plenoxels) trained from scratch on LLFF and TanksAndTemples datasets.}
\label{table-INGP2others}
\end{table*}
% The qualitative results(PSNR / SSIM / $\text{LPIPS}_{Alex}$ / $\text{LPIPS}_{VGG}$) of the conversion from INGP based Hash to other representations (VM-decomposition / MLP / Tensor)) on .

\subsubsection{Training and Distilling Details.} We implement our method with the PyTorch framework \cite{paszke2019pytorch} to train teachers and distill students. We use Adam Optimizer \cite{kingma2014adam} with initial learning rates of 0.02 and run 20k steps with batchsize of 4096 rays. For distilling, we initial the loss rate for volume-aligned, density, color and RGB with 2e-3, 2e-3, 2e-3 and 1 respectively. The first stage consumes 3k steps, the second stage consumes 5k steps, and the third stage will take all the rest steps. All the experiments are performed on a single NVIDIA V100 GPU. 
Please check the supplementary materials for more details.
% In particular, The Plenoxels does not use any MLP, so it is not suitable for use in multi-loss and block-wise when distilling, and we only study its convertibility
% TODO
%And we follow the implementation \cite{torch-ngp} for the sampling and empty space skipping

\subsection{Performance and Efficiency}
It should be noted that this is the first time to propose a conversion method between different representations, so we do not have any comparable baseline. Our experiments mainly focus on whether the conversion between different models can maintain the performance of the teacher or its own upper limit. And we also expect to get some benefits from the distillation between different structures.

\subsubsection{Quantitative Results}
For four representations (Hash / VM-decomposition / MLP / sparse tensors), we first train the models of each representation from scratch in 8 scenes on the NeRF-Synthetic dataset, and a total of 32 models are obtained as teachers. Then using the PVD proposed in this article to convert these teachers into the students with different structures. At the same time, we also consider the conversion between the same structures. We count the average metrics in Table \ref{table-mutual} after the conversion is complete. It can be seen that our method is very effective for the conversion. When a model is transformed into another forms, its performance has little difference with the result of training the model from scratch or the result of the teacher, which fully shows that the common representations based on radiance fields can be converted into each other. In addition, our PVD shows excellent nearly nondestructive performance in distillation between the same structures.
% It is worth noting in Table \ref{table-mutual} that the transformation of the same structure is the best way can maintain the performance of the original teacher.

In Fig.\ref{fig-heatmap}, we can see that the value of $\max(diff_{1}, diff_{2})$ is very close to 0, which means that the model obtained by distillation can be close to the teacher or training it from scratch. The performance of students is mainly limited by two aspects, one is the performance of teachers, and the other is the fitting ability of the student itself. Fig.\ref{fig-heatmap} shows strong evidence that our method has migrated knowledge from teacher to student to the maximum extent.
%we calculate the difference of PSNR between the model obtained by distillation and the teacher (represented by $diff_{tea}$), as well as the difference of PSNR between the the model obtained by distillation and the model itself trained from scratch (represented by $diff_{stu}$). And the $diff_{min}$ equals the minimum value between diff-tea and diff-self as shown in Fig.\ref{fig-heatmap}
% There are two situations that need to be explained: 1) The value of diff-tea is close to 0, but the value of diff-self is less than zero. 2) The value of diff-self is close to 0, but the value of diff-tea is less than 0. Scenario 1 shows that the performance of students is limited by the teacher. Scenario 2 shows that the performance of students is limited by the fitting ability of the student model itself. In either case, it shows that our method can fully exploit the upper limit fitting ability of students.
% , that is, the value of the student obtained by distillation minus the student trained from scratch.
% In addition, it is gratifying that almost half of the models can get better results than the training from scratch.
% It also should be noted that the above transformation does not use any real data. Only using the fake data generated by the teacher to train student is very important in the circumstance of less data or data privacy.
 %we further the benefits of different model transformations that can bring to the model. We further explore the benefits that transformations can bring to a model.

We further verify our method in Table \ref{table-INGP2others} on the LLFF and TanksAndTemples datasets. We use INGP as a teacher to distill NeRF, VM-decomposition and Plenoxels, and we compare them with the results obtained by training from scratch of these students. It can be seen from Table \ref{table-INGP2others} that our method is also effective on these two datasets. It is gratifying that the NeRF model obtained by our distillation performs better than its original implementation on TanksAndTemples dataset. This is mainly due to the fact that our PVD method provides more prior information to students, making training more efficient and fully improving the expression limit of the student.
\begin{table}[ht]
\begin{tabular}{cccc}
\hline
method & Lego                 & Orchids               & Truck                \\ \hline
NeRF   & 32.54/30h            & 20.36/35h             & 25.36/35h            \\
s-MLP  & 31.83/30min          & 20.61/100min          & 23.98/30min          \\
s-MLP  & \textbf{32.70}/1.5h & \textbf{21.25}/3h & \textbf{26.69}/1.7h \\ \hline
\end{tabular}
\caption{Comparison of running time. The teacher is based on the representation of VM-decomposition. We calculate the PSNR at different times for student and NeRF trained form scratch.}
\label{table-time}
\end{table}
 In addition to the possibility of improving the performance of the model, we also show another benefit from our method in Table \ref{table-time}. It can be clearly seen that our method obtains a NeRF model significantly faster than training the model from scratch. As we mentioned earlier, the process of distilling from a large NeRF model to a small NeRF model is abnormally inefficient, since it requires constantly querying the large NeRF model in both training and distilling. While our distillation between heterogeneous forms can achieve a more efficient distillation.
 %Tables 1 and 2 have fully demonstrated that the conversion between different structures is feasible. In addition to the possibility of improving performance,
% The performance of student exceeded that of training it from scratch on some data sets. We compare the performance of the original training implementation of NeRF and our distillation-based implementation in different training times. n this process, the large NeRF model needs to be queried all the time, so its efficiency is very low.  can be achieved use the characteristics of different structures to quickly guide the training of a small NeRF model, and the increase in efficiency is very significant.
%our method can obtain a NeRF model much faster than training it from scratch

\begin{figure}[ht]
\centering
\includegraphics[width=0.4\textwidth]{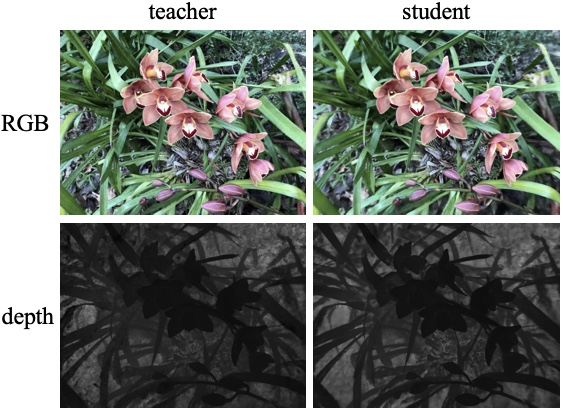} % Reduce the figure size so that it is slightly narrower than the column.
\caption{Qualitative comparison of depth in Orchids scene from LLFF dataset. The teacher is INGP and the student is s-MLP.}
\label{fig-llff-depth}
\end{figure}

\subsubsection{Qualitative Results} Fig.~\ref{fig-mutual} shows the qualitative results of mutual-conversion between Hash, VM-decomposition, MLP, and sparse tensors on NeRF-Synthetic dataset. We can see the excellent properties of PVD in maintaining the synthesis quality, as the visual quality of the student is often indistinguishably close to either the teacher or trained from scratch. We also show the result on TanksAndTemples dataset in Fig.\ref{fig-tank}. Our s-MLP achieves a better synthesis quality than NeRF training from scratch. The improvement is mainly due to the distillation between different structures. A powerful teacher can let the student approach its upper limit of expression capability. In addition, Fig.\ref{fig-llff-depth} shows that our method not only maintains the synthesis quality but also maintains the accuracy of the depth information of the scene.

%\subsubsection{Qualitative Results} Fig.~\ref{fig-mutual} shows the qualitative results of mutual-conversion between Hash, VM-decomposition, MLP, and sparse tensor representations on NeRF-Synthetic dataset. We can see the excellent properties of our PVD on maintaining the synthesis quality. The visual quality of student is infinitely close to the upper limit of its teacher or its own model, which proves that the scene knowledge learned by a model can be correctly transferred to others with different structure. We also show the result on TanksAndTemples dataset in Fig.\ref{fig-tank}. Our s-MLP achieves a better synthesis quality than NeRF training from scratch. The improvement is mainly due to the distillation between different structures. A powerful teacher can let the student approach its upper limit of expression ability infinitely. In addition, Fig.\ref{fig-llff-depth} shows that our method not only maintains the synthesis quality, but also maintains the depth accuracy of the scene.
% obtained from our distillation method with better synthesis quality

\begin{table}[h]
\resizebox{1.0\columnwidth}{!}{
\begin{tabular}{ccccc}
\hline
                  & PSNR$\uparrow$           & SSIM$\uparrow$           & $\text{LPIPS}_{Alex}\downarrow$          & $\text{LPIPS}_{VGG}\downarrow$          \\ \hline
w/o $\mathcal{L}_{2}^{v}$       & 29.63          & 0.937          & 0.065          & 0.087          \\
w/o $\mathcal{L}_{2}^{\sigma}$     & 30.01          & 0.939          & 0.063          & 0.084          \\
w/o $\mathcal{L}_{2}^{c}$     & 29.95          & 0.938          & 0.063          & 0.085          \\
w/o $\mathcal{L}_{2}^{rgb}$       & 27.07          & 0.908          & 0.945          & 0.116          \\
w/o sigma-constrain & 28.45          & 0.929          & 0.074          & 0.978          \\
w/o block-wise    & 29.62          & 0.941          & 0.060          & 0.082          \\
w/all             & \textbf{30.49} & \textbf{0.945} & \textbf{0.055} & \textbf{0.076} \\ \hline
\end{tabular}
}
\caption{An ablation study of our method. Metrics are averaged over the 8 scenes from  NeRF-Synthetic dataset in the conversion from VM-decomposition to s-MLP.}
\label{table-loss-ablation}
\end{table}

\subsection{Ablation studies and Limitations}
Our ablation studies demonstrate the degree of influence of each component in our method on the performance. We implement the conversion from VM-decomposition to MLP on the Synthetic-Nerf dataset as in Table \ref{table-loss-ablation}. It can be seen the intermediate feature loss we designed brings about 0.9dB PSNR improvement. It can also be seen that the performance will drop sharply without the restriction on the value of density. We also take the distillation without using block-wise strategy, and we find that it attains poor performance under the same budget of training time.
% Our ablation studies demonstrate the degree of influence of each component in our method on the performance. We implement the conversion from VM-decomposition decomposition to MLP on the Synthetic-NeRF dataset as in Table \ref{table-loss-ablation}. It can be seen the volume-aligned loss we designed brings about 0.9dB PSNR improvement. The performance will drop sharply without the restriction on the value of density. We also take the distillation without using block-wise strategy, and we find that it attains poor performance under the same time of using block-wise.

Our method also has some limitations inherited from the distillation. For example, the performance of student models is generally upper-bounded by the performances of teacher models, and in those cases further finetuning may be beneficial. Similarly, the modeling ability of the student model may limit its final performance. In addition, as both teacher and student models need be active during training, memory and computation cost will be duly increased. 
%Our method also relies on availability of accurate pose information, as in other NeRF methods.

\section{Conclusions}
In this work, we present PVD, a systematic distillation method that allows conversions between different NeRF architectures, including MLP, sparse tensor, low-rank tensor, and hash tables, while maintaining high sysnthesis quality. Central to the success of PVD is careful design of loss functions, a progressive distilling schemes utilizing intermediate volume representations, and special treatment of density values. By breaking through the barriers between different architectures, PVD allows downstream applications to optimally adapt the neural representation for the task at hand in a post hoc fashion. Empirical experiments solidly demonstrate the efficiency of our approach, on both synthetic and realworld datasets, both measured in quantitative PSNR and under visual inspection.
%In this work, we present PVD, a systematic distillation method that allows conversion between different NeRF architectures including MLP, sparse tensors, low-rank tensors, and hash tables, while maintaining high synthesis quality. Central to the success of PVD are careful design of loss functions, a progressive distilling schemes utilizing intermediate volume representations, and special treatment of density volumes. By breaking through the barriers between different architectures, PVD allows downstream applications to optimally adapt the neural representation for the task in hand in a post hoc fashion. Empirical experiments solidly demonstrate the efficiency of our approach, on both synthetic and realworld datasets.

\begin{center}
    \Large
     \textcolor[rgb]{0.,0.18,0.85}{\textbf{Supplementary Material}}
    \\[18pt]
    \normalsize 
\end{center}

\section{Overview}
In this supplementary material, we present additional details for our experiments, including the architecture of models, training details, further analysis and ablation studies of density, finetuning results and more detailed experimental results on each scene.

%This document details the formatting requirements for anonymous submissions. The requirements are the same as for camera ready papers but with a few notable differences:

\section{Model Settings}
In subsection Network Architecture, we have show several basic design in the models used in our method. A more detailed configuration of the structure used in this article is as follows:
\begin{itemize}
\item For Hash \cite{muller2022instant}, we set the coarsest resolution, the finest resolution, levels, hashtable size and feature dimensions to 16, $2048 \times \text{scene size}$, 14, $2^{19}$, and 2 respectively. The MLP followed the hashtables has two hidden layers with a width of 64, The activation function is ReLU for each hidden layer and sigmoid for the output layer of color.
\item For MLP, we are mainly based on the Pytorch implementation of the NeRF \cite{lin2020NeRFpytorch}. The whole network contains 8 FC layers with ReLU as the hidden activations and each hidden layer have 256 channels. The positional encoding of the input location is passed through this layers, An additional layer outputs the volume density (we do not use a ReLU to keep the value of density nonnegative, in our implementation, the value of density can be negative). A hidden layer would concatenate with the positional encoding of the input viewing direction followed by a FC layer with 128 channels. The activation function is also sigmoid on the output layer of color. Unlike the implementation in the original paper, we do not use an additional fine network, which is mainly to adapt to the distillation process, but we find that even without the fine network, an MLP-based network can still be obtained through distillation with high performance.
\item For VM-decomposition, We follow the implementation in \cite{chen2022tensorf}, and use a total of 48 components, where the resolution of the Matrix is 300$\times$300 for Synthetic-NeRF and TanksAndTemples datasets, and 640$\times$640 for LLFF dataset. We use three-order SH coefficients for the RGB channels. And a small MLP with two FC layers with 128-channel and ReLU activation used to interpret the density and color from the VM-decomposition features. We utilize an L1 norm loss and a TV loss on the Matrix and Vector factors.
\item For sparse tensors, we mainly refer to the implementation method of Plenoxels \cite{fridovich2022plenoxels}. But we do not using CUDA to accelerate the whole training process. A grid with the resolution 128$\times$128$\times$128 is used in the Synthetic-NeRF and TanksAndTemples datasets, and the resolution 512$\times$512$\times$128 is used in the LLFF datasets. The spherical harmonics is set to degree 2. Our task is not to get the best configuration for a structure, but to verify the conversion ability between different structures, so we just use a lower resolution as mentioned above in our experiments. And we do not implement the pruning of unnecessary voxels for simplicity.
\end{itemize}

\section{Implementation Details}
\subsection{Distillation Details}We use the Adam optimizer with initial learning rates of 0.02 (while 0.001 for the MLP decoder) in different structures. And we run 20k steps with batch size of 4096 rays. We initial the loss rate for volume-aligned, density, color and RGB with 2e-3, 2e-3, 2e-3 and 1 respectively. The first stage will be trained 3k steps, the second stage will be trained 5k, and the third stage will take all the rest steps. For VM-decomposition we do not implement the upsample strategy for the size of Matrix and Vector but just set them to a fixed resolution. And for sparse tensors, we follow the performing of trilinear color interpolation with a clip function to ensure the sample colors are always between 0 and 1. And a TV loss would be used in the VM-decomposition and sparse tensors with the rate of 1e-5 when distilling them, a L1 norm loss would be used in the VM-decomposition with the rate of 1e-4.

\subsection{Pseudo Data for Distillation}
As we mentioned in the section Experiments, the interconversion between different structures does not require any ground truth, and it is enough to generate pseudo-data using the pre-trained teacher model to provide the amount of data for the distillation process. We generate random poses from an orbit camera during each iteration.

\section{Analysis of Density in PVD}
We find that the loss of density $\sigma$ is hard to optimize directly. We attribute this problem to its specific numerical instability. When $\sigma$ is greater than or less than a certain value, its physical meaning is consistent (i.e., completely transparent or completely opaque). Therefore the value range of $\sigma$ can be too wide for a teacher, but in fact, only one interval of the density values play a key role.

\begin{table}[h]
\begin{tabular}{ccccl}
\hline
range                  & PSNR           & SSIM           & $\text{LPIPS}_{Alex}$    & $\text{LPIPS}_{Vgg}$     \\ \hline
{[}-31, 40{]} & 35.85          & 0.976          & 0.027          & 0.051          \\ \hline
{[}-20, 20{]}          & \textbf{35.85} & \textbf{0.976} & \textbf{0.027} & \textbf{0.051} \\
{[}-10, 10{]}          & \textbf{35.85} & \textbf{0.976} & \textbf{0.027} & \textbf{0.051} \\
{[}-2, 7{]}            & \textbf{35.82} & \textbf{0.976} & \textbf{0.027} & \textbf{0.051} \\
{[}-1, 7{]}            & 35.37          & 0.956          & 0.030          & 0.058          \\
{[}-2, 6{]}            & 34.92          & 0.937          & 0.038          & 0.062          \\ \hline
\end{tabular}
\caption{The influence of the density value range to a teacher trained in the scene hotdog from the Synthetic-NeRF dataset.}
\label{tab-clip-for-teacher}
\end{table}

% clipVSnoclip - loss density and RGB
\begin{figure}[ht]
    \centering
    \includegraphics[width=0.45\textwidth]{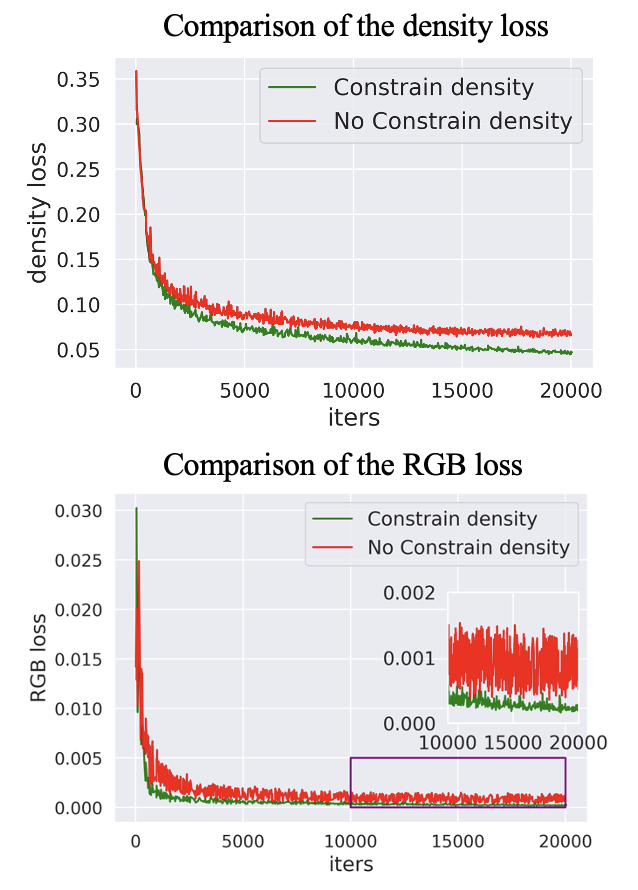} % Reduce the figure size so that it is slightly narrower than the column.
    \caption{Comparison of the density loss and RGB loss between constrained and unconstrained value range of the density on the hotdog scene from Synthetic-NeRF dataset. The constrained value range of density is [$-2, 7$]. }
    \label{fig-clip-vs-noclip-loss}
\end{figure}

In order to verify our conjecture, we designed the experiments shown in Table \ref{tab-clip-for-teacher}. In Table \ref{tab-clip-for-teacher}, we first train a teacher without constraint on the density value range in the hotdog scene, and then constrain the density value range with different intervals to observe its performance. It can be seen that there is almost no difference in performance outside a certain interval of density ([-2,7]). The same conclusion applies to other scenes.

Although the performance of models with different value range of density can be consistent, too large an interval of density is detrimental to the distillation task. When directly fitting the density, the student would tend to learn the larger values outside the interval of the density. Overfitting these big values makes the fit ability of the key interval ineffective, resulting in a decrease in student performance. As shown in Fig. \ref{fig-clip-vs-noclip-loss}, the loss of density with constrained is significantly lower than the loss of density with unconstrained, so that the RGB loss becomes lower and the performance of the student is effectively improved. We also provided detailed ablation results in Table \ref{tab-clip-vs-noclip-sys}, the performance of the student in various scenes has been greatly improved when constraining the range of density value.

\begin{table*}[b]
\centering
\resizebox{1.95\columnwidth}{!}{
\begin{tabular}{c|ccccccccc}
\hline
scene           & lego           & chair          & drum           & ficus          & hotdog         & materials      & mic            & ship           & Avg            \\ \hline
w/o PSNR        & 30.56          & 28.92          & 24.50          & 28.16          & 30.32          & 27.32          & 30.40          & 27.44          & 28.45          \\
w/ PSNR         & \textbf{32.84} & \textbf{31.59} & \textbf{25.01} & \textbf{29.94} & \textbf{35.11} & \textbf{29.01} & \textbf{31.97} & \textbf{28.46} & \textbf{30.49} \\ \hline
w/o SSIM        & 0.947          & 0.935          & 0.917          & 0.953          & 0.939          & 0.932          & 0.969          & 0.845          & 0.929          \\
w/ SSIM         & \textbf{0.963} & \textbf{0.956} & \textbf{0.926} & \textbf{0.967} & \textbf{0.970} & \textbf{0.948} & \textbf{0.976} & \textbf{0.860} & \textbf{0.945} \\ \hline
w/o $\text{LPIPS}_{Alex}$ & 0.031          & 0.077          & 0.087          & 0.037          & 0.087          & 0.054          & 0.036          & 0.183          & 0.074          \\
w/ $\text{LPIPS}_{Alex}$  & \textbf{0.022} & \textbf{0.051} & \textbf{0.078} & \textbf{0.024} & \textbf{0.039} & \textbf{0.043} & \textbf{0.028} & \textbf{0.158} & \textbf{0.055} \\ \hline
w/o $\text{LPIPS}_{Vgg}$  & 0.072          & 0.083          & 0.104          & 0.058          & 0.114          & 0.085          & 0.044          & 0.223          & 0.978          \\
w/ $\text{LPIPS}_{Vgg}$   & \textbf{0.053} & \textbf{0.061} & \textbf{0.095} & \textbf{0.041} & \textbf{0.060} & \textbf{0.070} & \textbf{0.032} & \textbf{0.203} & \textbf{0.076} \\ \hline
\end{tabular}
}
\caption{The metrics on Synthetic-NeRF dataset. "w/o" indicates without constraining the value range of density, and 'w/' means with constraining the value range of density  }
\label{tab-clip-vs-noclip-sys}
\end{table*}

\section{Selecting K for MLP}
In general, picking a proper K in Table 1 in the main article is striking a balance between performance and training time. We conducted an ablation study of distilling Hash into MLP on Synthetic-NeRF dataset and the average (PSNR, training time) are:
K=2(30.39,1.35h)
K=3(30.58,1.38h)
K=4(30.70,1.42h)
K=5(30.69,1.47h)
K=6(30.73,1.52h).
Here we get higher PSNR with larger K, which implies using more layers to fit hash tables can improve performance. In contrast, having smaller K reduces the training time due to our blockwise distillation strategy. In this case, K=4 would be a Pareto optimum.

\section{Finetuning Effects}
We divide the finetuning effects into two cases:

Case1: The teacher is superior in modeling capabilities.
Finetuning has little benefit in this case. An example PSNR results on Synthetic-NeRF dataset: Hash$\to $Tensors:27.85, finetune:27.82. The main reason is that a superior teacher can provide sufficient pseudo-datasets far beyond the number of real train datasets to train students adequately.

Case2: The student is superior in modeling capabilities.
In this case, the performance of the student is improved after finetuning because the student's performance is limited by the teacher when distilling. An example PSNR results are: Tensors$\to$Hash:27.32, finetune:32.41.

It should be noted that the primary role of our method is to exploit various properties of different structures as explained in A1. Hence it still makes sense to distill to a student with inferior modeling capabilities. Nevertheless, common tricks like increasing model parameters can be applied to better match teacher and student capabilities to avoid unnecessary loss of information.

\section{Per-scene Breakdown}
\begin{figure*}[t]
    \centering
    \includegraphics[width=0.95\textwidth]{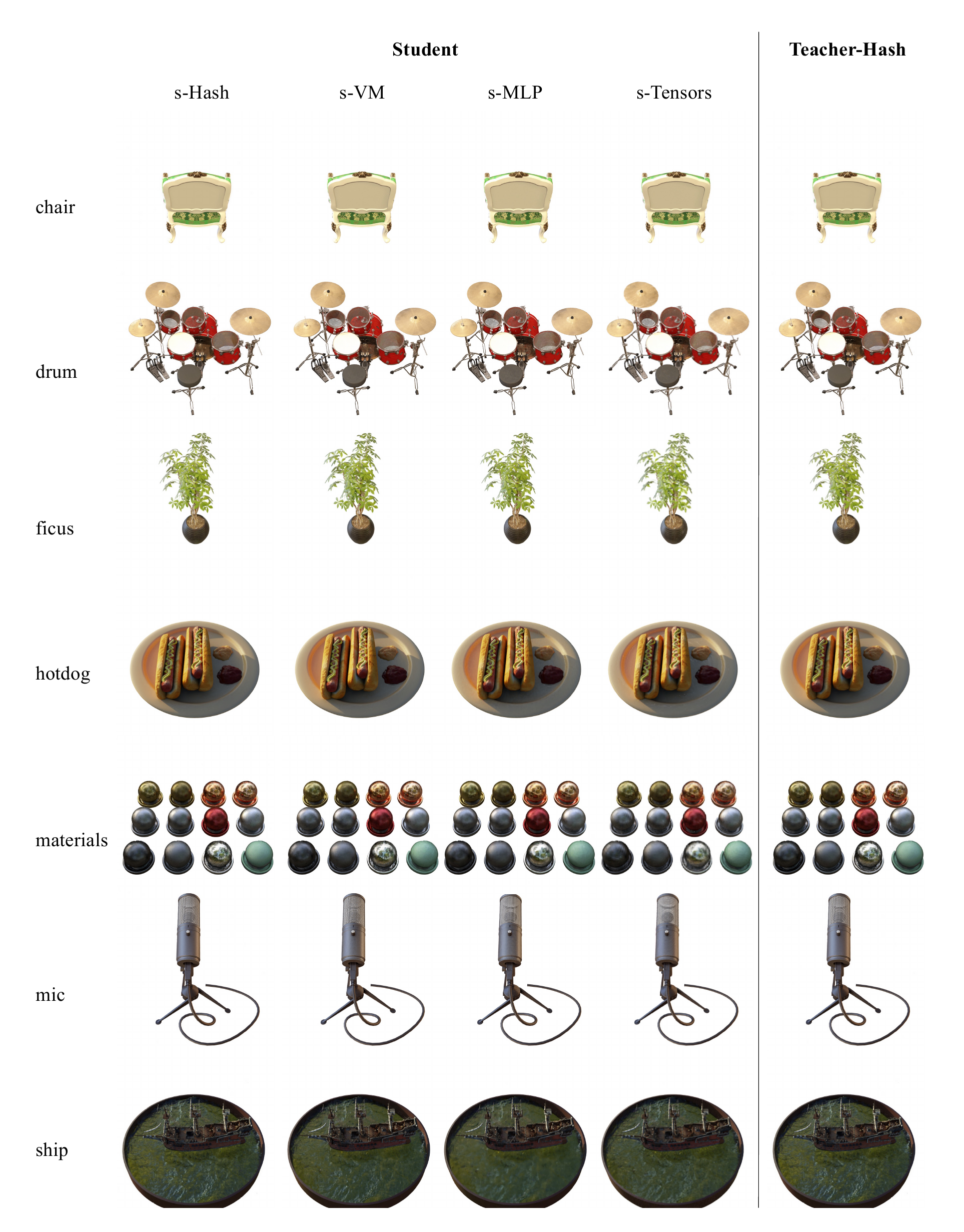} % Reduce the figure size so that it is slightly narrower than the column.
    \caption{Visual results of mutual-conversion on the Synthetic-NeRF dataset. Teacher is the model based on hash structure.}
    \label{fig-tea-hash-mutual-syn}
\end{figure*}

\begin{figure*}[t]
    \centering
    \includegraphics[width=0.95\textwidth]{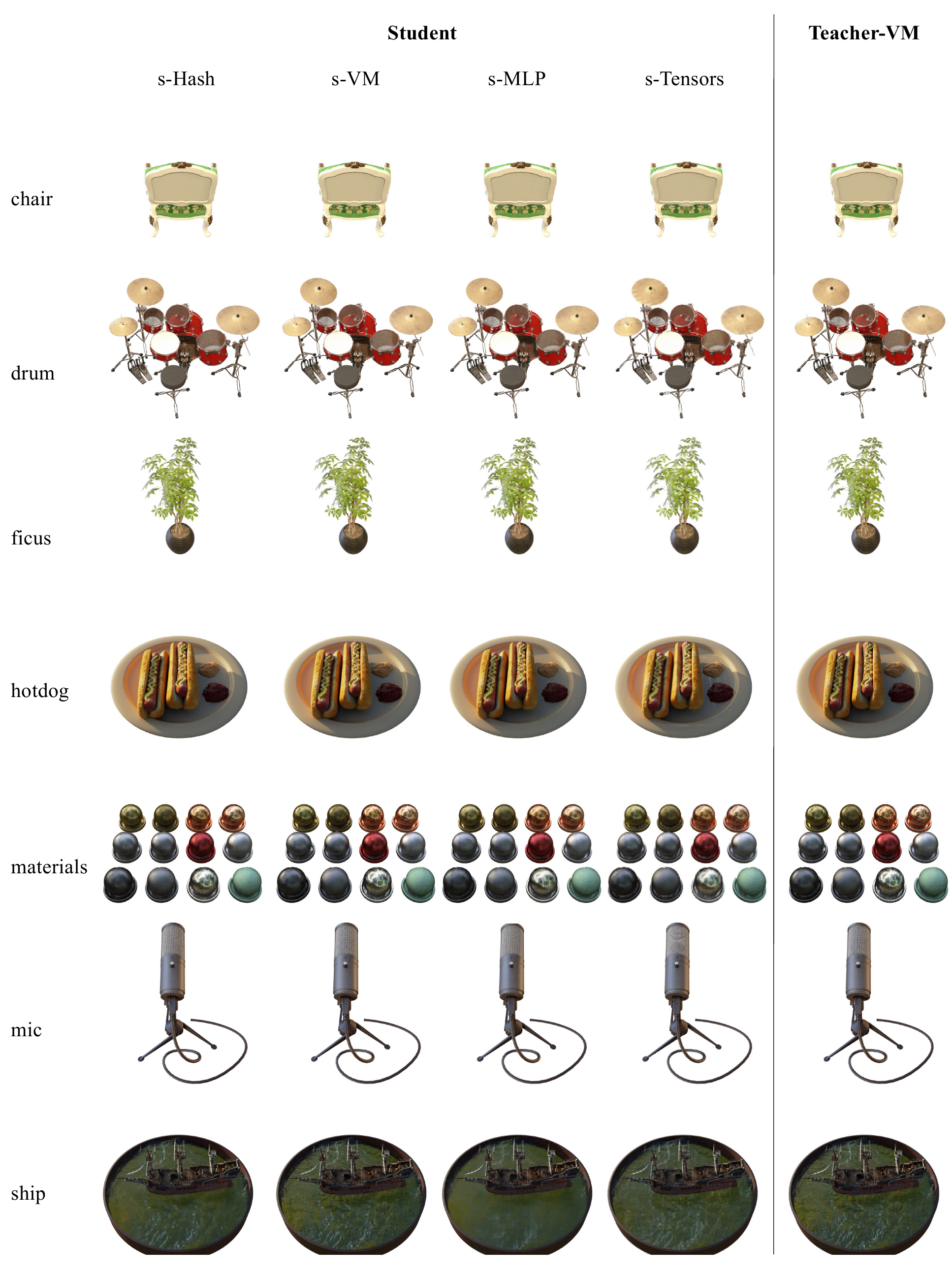} % Reduce the figure size so that it is slightly narrower than the column.
 \caption{Visual results of mutual-conversion on the Synthetic-NeRF dataset. Teacher is the model based on VM-composition structure.}
    \label{fig-tea-vm-mutual-syn}
\end{figure*}

\begin{figure*}[t]
    \centering
    \includegraphics[width=0.95\textwidth]{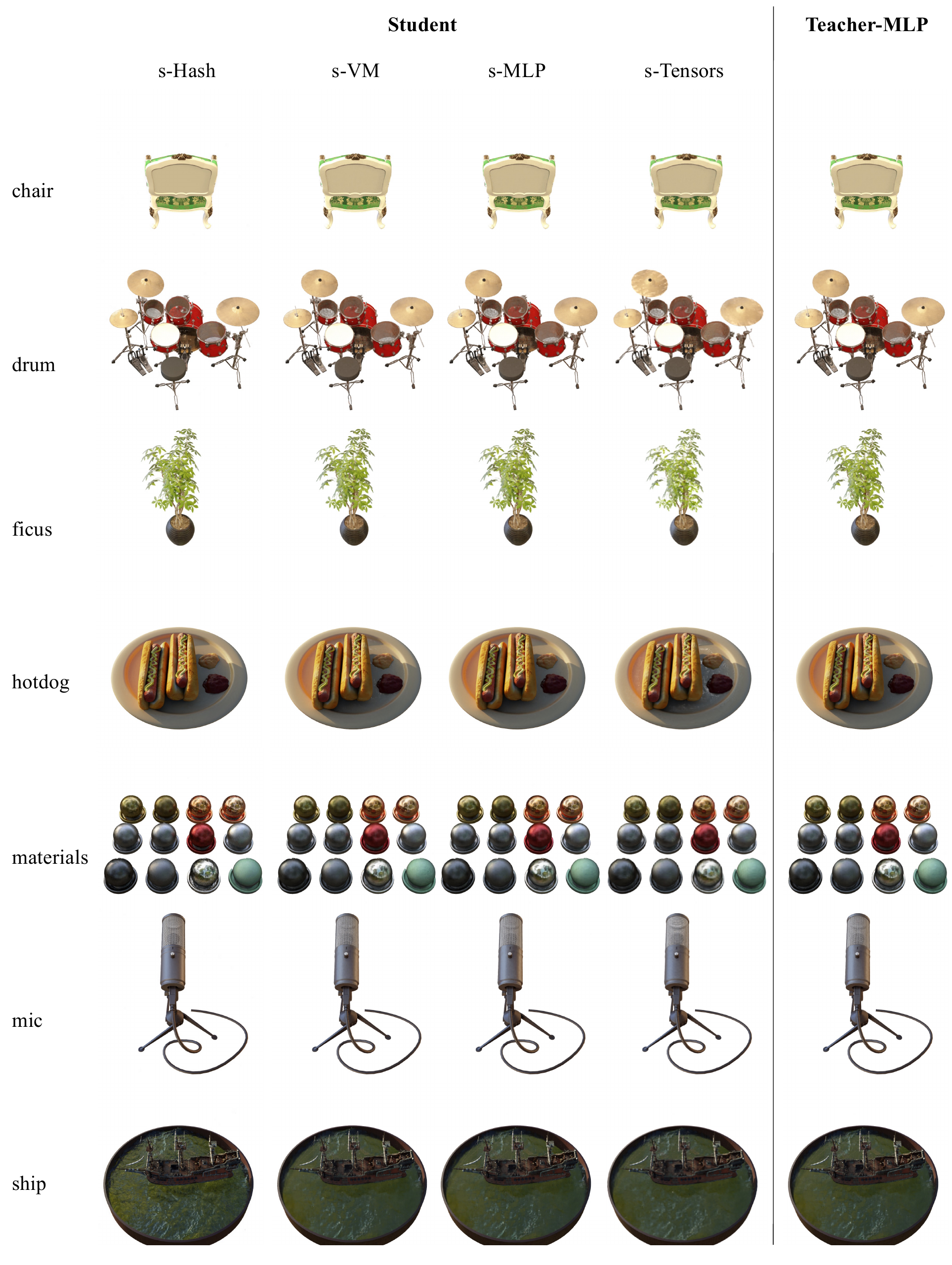} % Reduce the figure size so that it is slightly narrower than the column.
 \caption{Visual results of mutual-conversion on the Synthetic-NeRF dataset. Teacher is the model based on MLP structure.}
    \label{fig-tea-mlp-mutual-syn}
\end{figure*}

\begin{figure*}[t]
    \centering
    \includegraphics[width=0.95\textwidth]{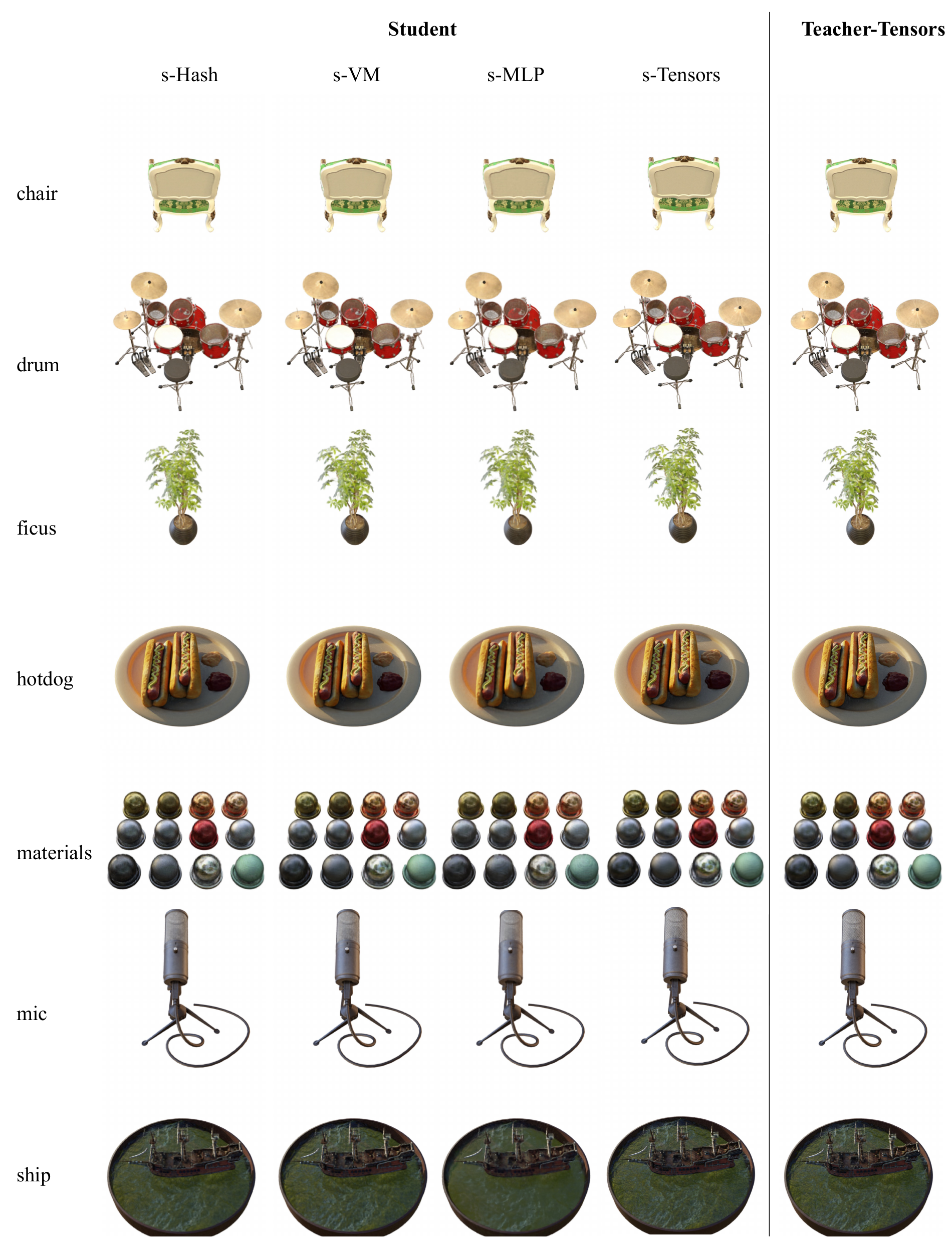} % Reduce the figure size so that it is slightly narrower than the column.
 \caption{Visual results of mutual-conversion on the Synthetic-NeRF dataset. Teacher is the model based on sparse tensors structure.}
    \label{fig-tea-tensor-mutual-syn}
\end{figure*}

\begin{figure*}[t]
    \centering
    \includegraphics[width=0.95\textwidth]{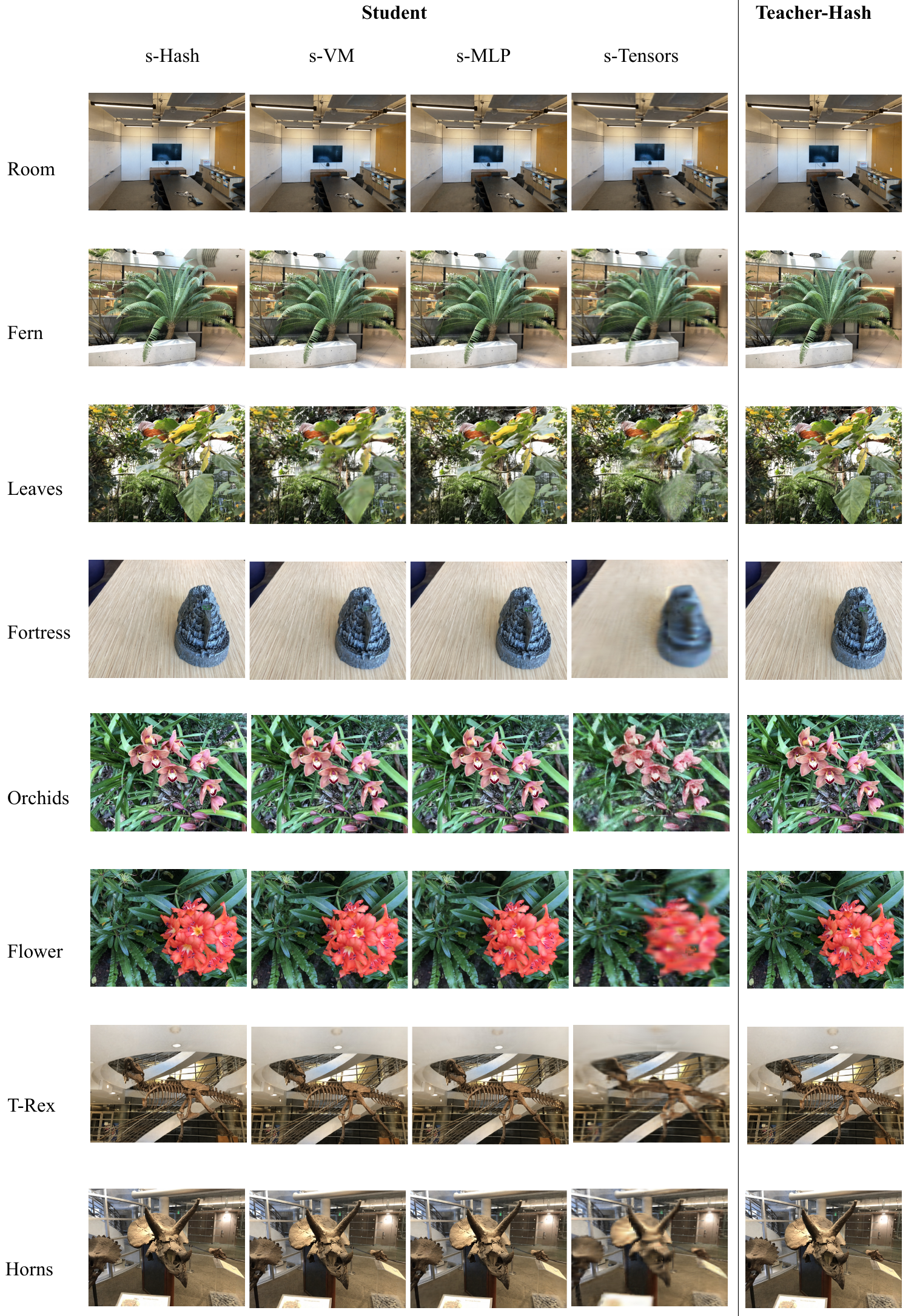} % Reduce the figure size so that it is slightly narrower than the column.
 \caption{Visual results of mutual-conversion on the LLFF dataset. Teacher is the model based on hash structure.}
    \label{fig-ngp2others-llff}
\end{figure*}

\begin{figure*}[t]
    \centering
    \includegraphics[width=1\textwidth]{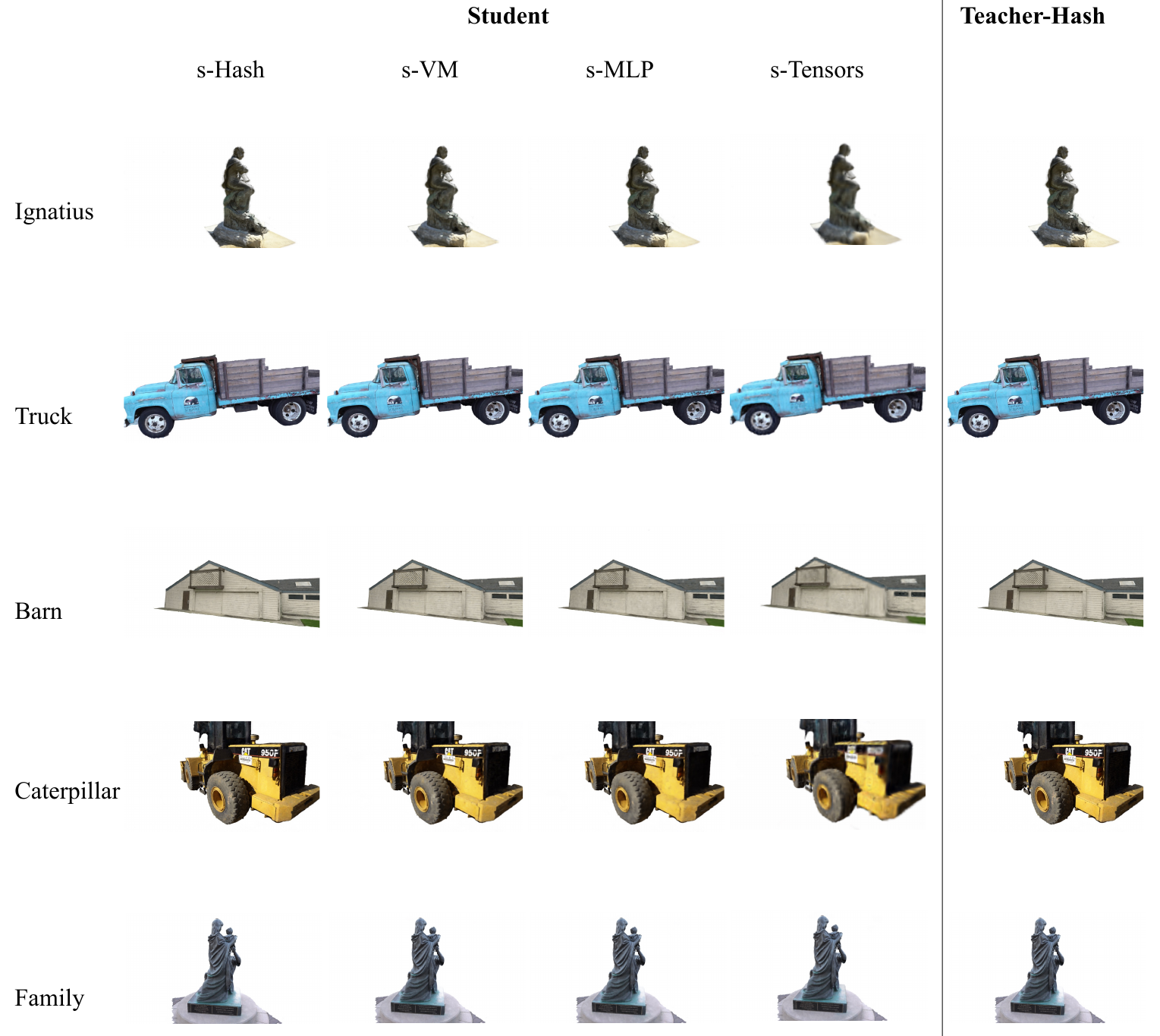} % Reduce the figure size so that it is slightly narrower than the column.
 \caption{Visual results of mutual-conversion on the TanksAndTemples dataset. Teacher is the model based on hash structure.}
    \label{fig-ngp2others-tanks}
\end{figure*}

% Table \ref{tab-ssim-mutual-syn}, Table \ref{tab-alex-mutual-syn},
% , Fig.\ref{fig-tea-vm-mutual-syn}, Fig.\ref{fig-tea-mlp-mutual-syn},
Table \ref{tab-psnr-mutual-syn} to  Table \ref{tab-vgg-mutual-syn} and Fig.\ref{fig-tea-hash-mutual-syn} to Fig. \ref{fig-tea-tensor-mutual-syn} give concrete quantitative and qualitative results of mutual-conversion between Hash / VM-decomposition / MLP / sparse tensors on the 7 scenes from the Synthetic-NeRF dataset.
Table \ref{tab-gp2others-llff} and Table \ref{tab-gp2others-tanks} show the quantitative results of models (s-VM, s-MLP, s-Tensors) obtained by our PVD with the models (TensoRF-VM, NeRF, Plenoxels) trained from scratch on LLFF and TanksAndTemples datasets. And The corresponding visual quality follows as Fig.\ref{fig-ngp2others-llff} and Fig.\ref{fig-ngp2others-tanks}.

\begin{table*}[t]
\centering
\resizebox{1.95\columnwidth}{!}{
\begin{tabular}{c|ccccccccc}
\hline
Models                      & Avg                     & lego                    & chair                   & drum                    & ficus                   & hotdog                  & materials               & mic                     & ship                    \\ \hline
\textit{\textbf{t-Hash}}    & \textit{\textbf{32.58}} & \textit{\textbf{35.70}} & \textit{\textbf{34.33}} & \textit{\textbf{25.91}} & \textit{\textbf{32.87}} & \textit{\textbf{36.60}} & \textit{\textbf{29.74}} & \textit{\textbf{35.36}} & \textit{\textbf{30.27}} \\ \hline
s-Hash                      & 32.58                   & 35.69                   & 34.32                   & 25.91                   & 32.81                   & 36.59                   & 29.71                   & 35.35                   & 30.27                   \\
s-VM                        & 31.33                   & 34.51                   & 32.45                   & 25.53                   & 31.08                   & 35.78                   & 29.20                   & 33.08                   & 29.03                   \\
s-MLP                       & 30.76                   & 32.70                   & 31.58                   & 25.29                   & 31.63                   & 34.92                   & 29.14                   & 32.68                   & 28.17                   \\
s-Tensors                   & 27.85                   & 28.59                   & 29.32                   & 23.66                   & 26.49                   & 32.94                   & 26.14                   & 29.50                   & 26.16                   \\ \hline
\textit{\textbf{t-VM}}      & \textit{\textbf{31.52}} & \textit{\textbf{34.72}} & \textit{\textbf{33.21}} & \textit{\textbf{25.61}} & \textit{\textbf{30.74}} & \textit{\textbf{35.85}} & \textit{\textbf{29.65}} & \textit{\textbf{32.94}} & \textit{\textbf{29.47}} \\ \hline
s-Hash                      & 30.96                   & 33.66                   & 32.58                   & 25.33                   & 30.52                   & 35.20                   & 28.83                   & 32.62                   & 28.90                   \\
s-VM                        & 31.52                   & 34.72                   & 33.21                   & 25.61                   & 30.74                   & 35.85                   & 29.65                   & 32.94                   & 29.47                   \\
s-MLP                       & 30.49                   & 32.84                   & 31.59                   & 25.01                   & 29.94                   & 35.11                   & 29.01                   & 31.97                   & 28.46                   \\
s-Tensors                   & 27.72                   & 28.27                   & 29.46                   & 23.54                   & 25.94                   & 32.64                   & 26.42                   & 29.37                   & 26.14                   \\ \hline
\textit{\textbf{t-MLP}}     & \textit{\textbf{30.78}} & \textit{\textbf{33.64}} & \textit{\textbf{31.88}} & \textit{\textbf{24.95}} & \textit{\textbf{30.21}} & \textit{\textbf{35.35}} & \textit{\textbf{29.05}} & \textit{\textbf{32.40}} & \textit{\textbf{28.80}} \\ \hline
s-Hash                      & 30.52                   & 32.89                   & 31.66                   & 24.88                   & 30.12                   & 35.08                   & 28.68                   & 32.31                   & 28.57                   \\
s-VM                        & 30.29                   & 32.91                   & 31.55                   & 24.82                   & 29.47                   & 35.00                   & 28.58                   & 31.79                   & 28.22                   \\
s-MLP                       & 30.78                   & 33.63                   & 31.88                   & 24.95                   & 30.21                   & 35.35                   & 29.04                   & 32.40                   & 28.80                   \\
s-Tensors                   & 27.44                   & 28.11                   & 29.40                   & 23.29                   & 25.22                   & 32.81                   & 25.73                   & 29.25                   & 25.75                   \\ \hline
\textit{\textbf{t-Tensors}} & \textit{\textbf{27.49}} & \textit{\textbf{27.90}} & \textit{\textbf{29.07}} & \textit{\textbf{23.34}} & \textit{\textbf{26.18}} & \textit{\textbf{32.23}} & \textit{\textbf{26.08}} & \textit{\textbf{29.29}} & \textit{\textbf{25.87}} \\ \hline
s-Hash                      & 27.32                   & 27.68                   & 28.94                   & 23.23                   & 26.13                   & 31.93                   & 25.78                   & 29.20                   & 25.71                   \\
s-VM                        & 27.46                   & 27.86                   & 29.04                   & 23.32                   & 26.20                   & 32.17                   & 26.00                   & 29.28                   & 25.84                   \\
s-MLP                       & 26.87                   & 27.02                   & 28.41                   & 22.96                   & 25.98                   & 31.20                   & 25.17                   & 28.97                   & 25.31                   \\
s-Tensors                   & 27.49                   & 27.89                   & 29.06                   & 23.34                   & 26.18                   & 32.22                   & 26.08                   & 29.29                   & 25.87                   \\ \hline
\end{tabular}
}
\caption{ The PSNR results of mutual-conversion between Hash / VM-decomposition / MLP /
sparse tensors representations on Synthetic-NeRF dataset. The bold italics number represents the teacher's metric, and the four numbers below represent the student's metrics obtained by distillation from the teacher. The s- means student.}
\label{tab-psnr-mutual-syn}
\end{table*}

% ssim-mutual
\begin{table*}[t]
\centering
\resizebox{1.85\columnwidth}{!}{
\begin{tabular}{c|ccccccccc}
\hline
Models                      & Avg                     & lego                    & chair                   & drum                    & ficus                   & hotdog                  & materials               & mic                     & ship                    \\ \hline
\textit{\textbf{t-Hash}}    & \textit{\textbf{0.960}} & \textit{\textbf{0.981}} & \textit{\textbf{0.981}} & \textit{\textbf{0.936}} & \textit{\textbf{0.982}} & \textit{\textbf{0.978}} & \textit{\textbf{0.950}} & \textit{\textbf{0.988}} & \textit{\textbf{0.886}} \\ \hline
s-Hash                      & 0.960                   & 0.981                   & 0.981                   & 0.936                   & 0.981                   & 0.978                   & 0.950                   & 0.988                   & 0.886                   \\
s-VM                        & 0.954                   & 0.976                   & 0.969                   & 0.934                   & 0.975                   & 0.974                   & 0.948                   & 0.982                   & 0.875                   \\
s-MLP                       & 0.946                   & 0.960                   & 0.959                   & 0.929                   & 0.975                   & 0.965                   & 0.948                   & 0.978                   & 0.856                   \\
s-Tensors                   & 0.921                   & 0.927                   & 0.929                   & 0.903                   & 0.938                   & 0.959                   & 0.921                   & 0.962                   & 0.831                   \\ \hline
\textit{\textbf{t-VM}}      & \textit{\textbf{0.955}} & \textit{\textbf{0.978}} & \textit{\textbf{0.971}} & \textit{\textbf{0.932}} & \textit{\textbf{0.973}} & \textit{\textbf{0.976}} & \textit{\textbf{0.951}} & \textit{\textbf{0.982}} & \textit{\textbf{0.883}} \\ \hline
s-Hash                      & 0.949                   & 0.969                   & 0.967                   & 0.929                   & 0.971                   & 0.972                   & 0.943                   & 0.980                   & 0.868                   \\
s-VM                        & 0.955                   & 0.978                   & 0.971                   & 0.932                   & 0.973                   & 0.976                   & 0.951                   & 0.982                   & 0.883                   \\
s-MLP                       & 0.945                   & 0.963                   & 0.956                   & 0.926                   & 0.967                   & 0.970                   & 0.948                   & 0.976                   & 0.860                   \\
s-Tensors                   & 0.921                   & 0.925                   & 0.929                   & 0.905                   & 0.935                   & 0.957                   & 0.924                   & 0.962                   & 0.835                   \\ \hline
\textit{\textbf{t-MLP}}     & \textit{\textbf{0.946}} & \textit{\textbf{0.968}} & \textit{\textbf{0.960}} & \textit{\textbf{0.925}} & \textit{\textbf{0.966}} & \textit{\textbf{0.970}} & \textit{\textbf{0.946}} & \textit{\textbf{0.977}} & \textit{\textbf{0.863}} \\ \hline
s-Hash                      & 0.944                   & 0.962                   & 0.958                   & 0.924                   & 0.966                   & 0.969                   & 0.942                   & 0.977                   & 0.859                   \\
s-VM                        & 0.944                   & 0.966                   & 0.957                   & 0.923                   & 0.962                   & 0.969                   & 0.942                   & 0.975                   & 0.857                   \\
s-MLP                       & 0.946                   & 0.968                   & 0.960                   & 0.925                   & 0.966                   & 0.970                   & 0.946                   & 0.977                   & 0.863                   \\
s-Tensors                   & 0.918                   & 0.924                   & 0.930                   & 0.902                   & 0.929                   & 0.958                   & 0.919                   & 0.962                   & 0.825                   \\ \hline
\textit{\textbf{t-Tensors}} & \textit{\textbf{0.917}} & \textit{\textbf{0.918}} & \textit{\textbf{0.927}} & \textit{\textbf{0.896}} & \textit{\textbf{0.931}} & \textit{\textbf{0.956}} & \textit{\textbf{0.924}} & \textit{\textbf{0.961}} & \textit{\textbf{0.830}} \\ \hline
s-Hash                      & 0.913                   & 0.912                   & 0.924                   & 0.893                   & 0.930                   & 0.952                   & 0.911                   & 0.959                   & 0.823                   \\
s-VM                        & 0.916                   & 0.917                   & 0.926                   & 0.895                   & 0.931                   & 0.955                   & 0.918                   & 0.960                   & 0.828                   \\
s-MLP                       & 0.906                   & 0.898                   & 0.918                   & 0.888                   & 0.927                   & 0.948                   & 0.904                   & 0.957                   & 0.811                   \\
s-Tensors                   & 0.917                   & 0.918                   & 0.926                   & 0.896                   & 0.931                   & 0.956                   & 0.920                   & 0.961                   & 0.830                   \\ \hline
\end{tabular}
}
\caption{ The SSIM results of mutual-conversion between Hash / VM-decomposition / MLP /
sparse tensors representations on  Synthetic-NeRF dataset. The bold italics number represents the teacher's metric, and the four numbers below represent the student's metrics obtained by distillation from the teacher. The s- means
student.}
\label{tab-ssim-mutual-syn}
\end{table*}

% lpips-alex
\begin{table*}[t]
\centering
\resizebox{1.85\columnwidth}{!}{
\begin{tabular}{c|ccccccccc}
\hline
Models                      & Avg                     & lego                    & chair                   & drum                    & ficus                   & hotdog                  & materials               & mic                     & ship                    \\ \hline
\textit{\textbf{t-Hash}}    & \textit{\textbf{0.032}} & \textit{\textbf{0.009}} & \textit{\textbf{0.014}} & \textit{\textbf{0.058}} & \textit{\textbf{0.018}} & \textit{\textbf{0.021}} & \textit{\textbf{0.036}} & \textit{\textbf{0.010}} & \textit{\textbf{0.091}} \\ \hline
s-Hash                      & 0.032                   & 0.009                   & 0.014                   & 0.058                   & 0.018                   & 0.021                   & 0.036                   & 0.010                   & 0.092                   \\
s-VM                        & 0.042                   & 0.012                   & 0.028                   & 0.065                   & 0.025                   & 0.029                   & 0.041                   & 0.018                   & 0.125                   \\
s-MLP                       & 0.056                   & 0.024                   & 0.052                   & 0.074                   & 0.023                   & 0.049                   & 0.037                   & 0.025                   & 0.165                   \\
s-Tensors                   & 0.100                   & 0.076                   & 0.099                   & 0.128                   & 0.050                   & 0.071                   & 0.102                   & 0.060                   & 0.220                   \\ \hline
\textit{\textbf{t-VM}}      & \textit{\textbf{0.040}} & \textit{\textbf{0.012}} & \textit{\textbf{0.024}} & \textit{\textbf{0.066}} & \textit{\textbf{0.023}} & \textit{\textbf{0.027}} & \textit{\textbf{0.040}} & \textit{\textbf{0.019}} & \textit{\textbf{0.111}} \\ \hline
s-Hash                      & 0.047                   & 0.017                   & 0.030                   & 0.071                   & 0.024                   & 0.032                   & 0.047                   & 0.020                   & 0.141                   \\
s-VM                        & 0.040                   & 0.012                   & 0.024                   & 0.066                   & 0.023                   & 0.027                   & 0.040                   & 0.019                   & 0.111                   \\
s-MLP                       & 0.055                   & 0.022                   & 0.051                   & 0.078                   & 0.024                   & 0.039                   & 0.043                   & 0.028                   & 0.158                   \\
s-Tensors                   & 0.099                   & 0.077                   & 0.099                   & 0.125                   & 0.052                   & 0.072                   & 0.097                   & 0.062                   & 0.211                   \\ \hline
\textit{\textbf{t-MLP}}     & \textit{\textbf{0.049}} & \textit{\textbf{0.016}} & \textit{\textbf{0.039}} & \textit{\textbf{0.074}} & \textit{\textbf{0.027}} & \textit{\textbf{0.034}} & \textit{\textbf{0.037}} & \textit{\textbf{0.027}} & \textit{\textbf{0.139}} \\ \hline
s-Hash                      & 0.053                   & 0.021                   & 0.045                   & 0.077                   & 0.027                   & 0.037                   & 0.042                   & 0.027                   & 0.155                   \\
s-VM                        & 0.056                   & 0.020                   & 0.047                   & 0.077                   & 0.035                   & 0.039                   & 0.047                   & 0.030                   & 0.156                   \\
s-MLP                       & 0.049                   & 0.016                   & 0.039                   & 0.074                   & 0.027                   & 0.034                   & 0.037                   & 0.027                   & 0.139                   \\
s-Tensors                   & 0.098                   & 0.071                   & 0.096                   & 0.121                   & 0.055                   & 0.071                   & 0.094                   & 0.053                   & 0.223                   \\ \hline
\textit{\textbf{t-Tensors}} & \textit{\textbf{0.122}} & \textit{\textbf{0.106}} & \textit{\textbf{0.115}} & \textit{\textbf{0.155}} & \textit{\textbf{0.065}} & \textit{\textbf{0.087}} & \textit{\textbf{0.129}} & \textit{\textbf{0.077}} & \textit{\textbf{0.242}} \\ \hline
s-Hash                      & 0.119                   & 0.102                   & 0.109                   & 0.152                   & 0.064                   & 0.086                   & 0.130                   & 0.076                   & 0.234                   \\
s-VM                        & 0.121                   & 0.106                   & 0.115                   & 0.155                   & 0.065                   & 0.087                   & 0.130                   & 0.076                   & 0.241                   \\
s-MLP                       & 0.127                   & 0.111                   & 0.117                   & 0.158                   & 0.067                   & 0.096                   & 0.134                   & 0.078                   & 0.260                   \\
s-Tensors                   & 0.122                   & 0.106                   & 0.116                   & 0.156                   & 0.065                   & 0.087                   & 0.129                   & 0.077                   & 0.242                   \\ \hline
\end{tabular}
}
\caption{ The $\text{LPIPS}_{Alex}$ results of mutual-conversion between Hash / VM-decomposition / MLP /
sparse tensors representations on  Synthetic-NeRF dataset. The bold italics number represents the metric of the teacher, and the four numbers below it represent the metrics of the student obtained by distillation from the teacher. The s- means
student.}
\label{tab-alex-mutual-syn}
\end{table*}

%lpips-vgg
\begin{table*}[t]
\centering
\resizebox{2.0\columnwidth}{!}{
\begin{tabular}{c|ccccccccc}
\hline
Models                      & Avg                     & lego                    & chair                   & drum                    & ficus                   & hotdog                  & materials               & mic                     & ship                    \\ \hline
\textit{\textbf{t-Hash}}    & \textit{\textbf{0.055}} & \textit{\textbf{0.022}} & \textit{\textbf{0.031}} & \textit{\textbf{0.080}} & \textit{\textbf{0.026}} & \textit{\textbf{0.052}} & \textit{\textbf{0.070}} & \textit{\textbf{0.019}} & \textit{\textbf{0.144}} \\ \hline
s-Hash                      & 0.055                   & 0.022                   & 0.031                   & 0.080                   & 0.027                   & 0.052                   & 0.070                   & 0.019                   & 0.144                   \\
s-VM                        & 0.063                   & 0.029                   & 0.040                   & 0.080                   & 0.035                   & 0.054                   & 0.073                   & 0.023                   & 0.170                   \\
s-MLP                       & 0.076                   & 0.059                   & 0.055                   & 0.091                   & 0.034                   & 0.073                   & 0.068                   & 0.030                   & 0.205                   \\
s-Tensors                   & 0.106                   & 0.103                   & 0.086                   & 0.123                   & 0.070                   & 0.082                   & 0.105                   & 0.052                   & 0.228                   \\ \hline
\textit{\textbf{t-VM}}      & \textit{\textbf{0.061}} & \textit{\textbf{0.027}} & \textit{\textbf{0.040}} & \textit{\textbf{0.084}} & \textit{\textbf{0.035}} & \textit{\textbf{0.051}} & \textit{\textbf{0.071}} & \textit{\textbf{0.027}} & \textit{\textbf{0.160}} \\ \hline
s-Hash                      & 0.071                   & 0.044                   & 0.047                   & 0.089                   & 0.037                   & 0.058                   & 0.081                   & 0.028                   & 0.186                   \\
s-VM                        & 0.061                   & 0.027                   & 0.040                   & 0.084                   & 0.035                   & 0.051                   & 0.071                   & 0.027                   & 0.160                   \\
s-MLP                       & 0.076                   & 0.053                   & 0.061                   & 0.095                   & 0.041                   & 0.060                   & 0.070                   & 0.032                   & 0.203                   \\
s-Tensors                   & 0.107                   & 0.104                   & 0.088                   & 0.124                   & 0.074                   & 0.086                   & 0.103                   & 0.055                   & 0.223                   \\ \hline
\textit{\textbf{t-MLP}}     & \textit{\textbf{0.075}} & \textit{\textbf{0.044}} & \textit{\textbf{0.059}} & \textit{\textbf{0.096}} & \textit{\textbf{0.047}} & \textit{\textbf{0.061}} & \textit{\textbf{0.071}} & \textit{\textbf{0.036}} & \textit{\textbf{0.189}} \\ \hline
s-Hash                      & 0.079                   & 0.056                   & 0.061                   & 0.098                   & 0.046                   & 0.065                   & 0.077                   & 0.034                   & 0.201                   \\
s-VM                        & 0.079                   & 0.047                   & 0.061                   & 0.097                   & 0.051                   & 0.063                   & 0.080                   & 0.036                   & 0.201                   \\
s-MLP                       & 0.075                   & 0.044                   & 0.059                   & 0.096                   & 0.047                   & 0.061                   & 0.071                   & 0.036                   & 0.189                   \\
s-Tensors                   & 0.108                   & 0.103                   & 0.087                   & 0.126                   & 0.077                   & 0.084                   & 0.104                   & 0.050                   & 0.240                   \\ \hline
\textit{\textbf{t-Tensors}} & \textit{\textbf{0.112}} & \textit{\textbf{0.119}} & \textit{\textbf{0.093}} & \textit{\textbf{0.130}} & \textit{\textbf{0.078}} & \textit{\textbf{0.087}} & \textit{\textbf{0.103}} & \textit{\textbf{0.058}} & \textit{\textbf{0.230}} \\ \hline
s-Hash                      & 0.118                   & 0.125                   & 0.096                   & 0.134                   & 0.080                   & 0.093                   & 0.120                   & 0.061                   & 0.236                   \\
s-VM                        & 0.114                   & 0.120                   & 0.094                   & 0.130                   & 0.078                   & 0.088                   & 0.111                   & 0.058                   & 0.234                   \\
s-MLP                       & 0.125                   & 0.136                   & 0.102                   & 0.139                   & 0.083                   & 0.100                   & 0.122                   & 0.063                   & 0.262                   \\
s-Tensors                   & 0.113                   & 0.119                   & 0.093                   & 0.130                   & 0.078                   & 0.087                   & 0.108                   & 0.058                   & 0.231                   \\ \hline
\end{tabular}
}
\caption{ The $\text{LPIPS}_{Vgg}$ results of mutual-conversion between Hash / VM-decomposition / MLP / sparse tensors representations on  Synthetic-NeRF dataset. The bold italics number represents the metric of the teacher, and the four numbers below it represent the metrics of the student obtained by distillation from the teacher. The s- means
student.}
\label{tab-vgg-mutual-syn}
\end{table*}

% llff-ngp2others
\begin{table*}[t]
\centering
\resizebox{2.0\columnwidth}{!}{
\begin{tabular}{c|ccccccccc}
\hline
Models                   & Avg                     & Room                    & Fern                    & Leaves                  & Fortress                & Orchids                 & Flower                  & T-Rex                   & Horns                   \\ \hline
                         & \multicolumn{9}{c}{PSNR}                                                                                                                                                                                                                \\ \hline
\textit{\textbf{t-Hash}} & \textit{\textbf{26.70}} & \textit{\textbf{31.92}} & \textit{\textbf{26.19}} & \textit{\textbf{21.38}} & \textit{\textbf{30.48}} & \textit{\textbf{21.73}} & \textit{\textbf{27.17}} & \textit{\textbf{26.84}} & \textit{\textbf{27.86}} \\
TensoRF-VM               & \textbf{26.51}          & \textbf{31.80}          & \textbf{25.31}          & \textbf{21.34}          & \textbf{31.14}          & \textbf{20.02}          & \textbf{28.22}          & \textbf{26.61}          & \textbf{27.64}          \\
s-VM                     & 25.73                   & 30.69                   & 25.27                   & 20.22                   & 28.48                   & \textbf{20.83}          & 27.19                   & 26.30                   & 26.83                   \\
NeRF                     & \textbf{26.50}          & \textbf{32.70}          & 25.17                   & \textbf{20.92}          & \textbf{31.16}          & 20.36                   & \textbf{27.40}          & \textbf{26.80}          & \textbf{27.45}          \\
s-MLP                    & 25.77                   & 30.52                   & \textbf{25.19}          & 19.85                   & 30.10                   & \textbf{21.25}          & 26.51                   & 25.82                   & 26.95                   \\
Plenoxels128             & \textbf{21.69}          & \textbf{27.96}          & \textbf{22.17}          & \textbf{18.85}          & \textbf{23.30}          & \textbf{17.32}          & \textbf{21.31}          & \textbf{20.83}          & \textbf{21.83}          \\
s-Tensors                & 21.36                   & 27.20                   & 22.10                   & 18.06                   & 23.18                   & \textbf{18.02}          & 20.62                   & 20.38                   & 21.32                   \\ \hline
                         & \multicolumn{9}{c}{SSIM}                                                                                                                                                                                                                \\ \hline
\textit{\textbf{t-Hash}} & \textit{\textbf{0.832}} & \textit{\textbf{0.943}} & \textit{\textbf{0.835}} & \textit{\textbf{0.714}} & \textit{\textbf{0.883}} & \textit{\textbf{0.691}} & \textit{\textbf{0.833}} & \textit{\textbf{0.881}} & \textit{\textbf{0.876}} \\
TensoRF-VM               & \textbf{0.811}          & \textbf{0.948}          & 0.792                   & \textbf{0.690}          & \textbf{0.881}          & \textbf{0.641}          & 0.827                   & \textbf{0.880}          & \textbf{0.828}          \\
s-VM                     & 0.793                   & 0.931                   & \textbf{0.815}          & 0.629                   & 0.802                   & 0.635                   & \textbf{0.835}          & 0.871                   & 0.824                   \\
NeRF                     & \textbf{0.811}          & \textbf{0.948}          & \textbf{0.792}          & \textbf{0.690}          & \textbf{0.881}          & 0.641                   & \textbf{0.827}          & \textbf{0.880}          & \textbf{0.828}          \\
s-MLP                    & 0.784                   & 0.924                   & 0.783                   & 0.607                   & 0.862                   & \textbf{0.651}          & 0.781                   & 0.852                   & 0.812                   \\
Plenoxels128             & \textbf{0.607}          & \textbf{0.882}          & \textbf{0.661}          & \textbf{0.463}          & \textbf{0.562}          & 0.428                   & \textbf{0.614}          & \textbf{0.644}          & \textbf{0.609}          \\
s-Tensors                & 0.600                   & 0.867                   & 0.644                   & 0.452                   & 0.560                   & \textbf{0.475}          & 0.586                   & 0.625                   & 0.591                   \\ \hline
                         & \multicolumn{9}{c}{$\text{LPIPS}_{Alex}$}                                                                                                                                                                                                    \\ \hline
\textit{\textbf{t-Hash}} & \textit{\textbf{0.130}} & \textit{\textbf{0.094}} & \textit{\textbf{0.129}} & \textit{\textbf{0.213}} & \textit{\textbf{0.077}} & \textit{\textbf{0.204}} & \textit{\textbf{0.109}} & \textit{\textbf{0.115}} & \textit{\textbf{0.105}} \\
TensoRF-VM               & \textbf{0.135}          & \textbf{0.093}          & \textbf{0.161}          & \textbf{0.167}          & \textbf{0.084}          & \textbf{0.204}          & \textbf{0.121}          & \textbf{0.108}          & \textbf{0.146}          \\
s-VM                     & 0.195                   & 0.128                   & 0.177                   & 0.310                   & 0.183                   & 0.290                   & 0.144                   & 0.142                   & 0.191                   \\
NeRF                     & -                       & -                       & -                       & -                       & -                       & -                       & -                       & -                       & -                       \\
s-MLP                    & 0.213                   & 0.157                   & 0.241                   & 0.334                   & 0.123                   & 0.280                   & 0.197                   & 0.167                   & 0.207                   \\
Plenoxels128             & \textbf{0.527}          & \textbf{0.260}          & \textbf{0.473}          & \textbf{0.591}          & \textbf{0.644}          & 0.673                   & \textbf{0.531}          & \textbf{0.530}          & \textbf{0.521}          \\
s-Tensors                & 0.561                   & 0.316                   & 0.533                   & 0.608                   & 0.703                   & \textbf{0.549}          & 0.607                   & 0.595                   & 0.580                   \\ \hline
                         & \multicolumn{9}{c}{$\text{LPIPS}_{Vgg}$}                                                                                                                                                                                                     \\ \hline
\textit{\textbf{t-Hash}} & \textit{\textbf{0.231}} & \textit{\textbf{0.212}} & \textit{\textbf{0.227}} & \textit{\textbf{0.318}} & \textit{\textbf{0.164}} & \textit{\textbf{0.295}} & \textit{\textbf{0.195}} & \textit{\textbf{0.257}} & \textit{\textbf{0.187}} \\
TensoRF-VM               & \textbf{0.217}          & 0.181                   & \textbf{0.237}          & \textbf{0.230}          & \textbf{0.159}          & \textbf{0.283}          & \textbf{0.187}          & \textbf{0.236}          & \textbf{0.221}          \\
s-VM                     & 0.269                   & \textbf{0.128}          & 0.255                   & 0.366                   & 0.289                   & 0.349                   & 0.220                   & 0.274                   & 0.272                   \\
NeRF                     & \textbf{0.250}          & \textbf{0.178}          & \textbf{0.280}          & \textbf{0.316}          & \textbf{0.171}          & \textbf{0.321}          & \textbf{0.219}          & \textbf{0.249}          & \textbf{0.268}          \\
s-MLP                    & 0.310                   & 0.266                   & 0.323                   & 0.420                   & 0.234                   & 0.340                   & 0.294                   & 0.307                   & 0.303                   \\
Plenoxels128             & \textbf{0.52}           & \textbf{0.314}          & \textbf{0.429}          & 0.695                   & \textbf{0.577}          & \textbf{0.615}          & \textbf{0.502}          & \textbf{0.511}          & \textbf{0.517}          \\
s-Tensors                & 0.524                   & 0.346                   & 0.461                   & \textbf{0.617}          & 0.614                   & 0.598                   & 0.545                   & 0.546                   & 0.549                   \\ \hline
\end{tabular}
}
\caption{Comparing the per-scene metrics of models (s-VM, s-MLP, s-Tensors) obtained by PVD with the models (TensoRF-VM, NeRF, Plenoxels) trained from scratch on LLFF dataset.}
\label{tab-gp2others-llff}
\end{table*}

% Figure mutual-conversion in synthetic
\begin{table*}[t]
\centering
\resizebox{1.5\columnwidth}{!}{
% \usepackage{multirow}
\begin{tabular}{c|cccccc}
\hline
\multirow{2}{*}{Models}  & Avg                     & Ignatius                & Truck                   & Barn                    & Caterpillar             & Family                  \\ \cline{2-7} 
                         & \multicolumn{6}{c}{PSNR}                                                                                                                                  \\ \hline
\textit{\textbf{t-Hash}} & \textit{\textbf{29.26}} & \textit{\textbf{28.26}} & \textit{\textbf{28.54}} & \textit{\textbf{28.54}} & \textit{\textbf{26.47}} & \textit{\textbf{34.48}} \\
TensoRF-VM               & \textbf{28.06}          & \textbf{28.22}          & 26.81                   & \textbf{26.70}          & \textbf{25.43}          & \textbf{33.12}          \\
s-VM                     & 27.86                   & 27.44                   & \textbf{27.18}          & 26.33                   & 25.30                   & 33.07                   \\
NeRF                     & 25.78                   & 25.43                   & 25.36                   & 24.05                   & 23.75                   & 30.29                   \\
s-MLP                    & \textbf{27.50}          & \textbf{27.91}          & \textbf{26.69}          & \textbf{25.56}          & \textbf{24.93}          & \textbf{32.42}          \\
Plenoxels128             & 25.18                   & 25.42                   & \textbf{24.39}          & 23.44                   & \textbf{22.24}          & 30.41                   \\
s-Tensors                & \textbf{25.31}          & \textbf{25.53}          & 24.34                   & \textbf{23.58}          & 22.23                   & \textbf{30.86}          \\ \hline
                         & \multicolumn{6}{c}{SSIM}                                                                                                                                  \\ \hline
\textit{\textbf{t-Hash}} & \textit{\textbf{0.915}} & \textit{\textbf{0.948}} & \textit{\textbf{0.924}} & \textit{\textbf{0.871}} & \textit{\textbf{0.909}} & \textit{\textbf{0.969}} \\
TensoRF-VM               & \textbf{0.909}          & \textbf{0.943}          & \textbf{0.902}          & \textbf{0.845}          & \textbf{0.899}          & \textbf{0.957}          \\
s-VM                     & 0.899                   & 0.932                   & 0.898                   & 0.815                   & 0.894                   & 0.956                   \\
NeRF                     & 0.864                   & 0.920                   & 0.860                   & 0.750                   & 0.860                   & 0.932                   \\
s-MLP                    & \textbf{0.891}          & \textbf{0.938}          & \textbf{0.886}          & \textbf{0.801}          & \textbf{0.886}          & \textbf{0.948}          \\
Plenoxels128             & 0.865                   & \textbf{0.918}          & \textbf{0.850}          & 0.772                   & \textbf{0.851}          & 0.934                   \\
s-Tensors                & \textbf{0.866}          & \textbf{0.918}          & \textbf{0.850}          & \textbf{0.774}          & \textbf{0.851}          & \textbf{0.938}          \\ \hline
                         & \multicolumn{6}{c}{$\text{LPIPS}_{Alex}$}                                                                                                                            \\ \hline
\textit{\textbf{t-Hash}} & \textit{\textbf{0.106}} & \textit{\textbf{0.074}} & \textit{\textbf{0.096}} & \textit{\textbf{0.201}} & \textit{\textbf{0.122}} & \textit{\textbf{0.037}} \\
TensoRF-VM               & \textbf{0.145}          & \textbf{0.089}          & \textbf{0.145}          & \textbf{0.266}          & \textbf{0.161}          & \textbf{0.066}          \\
s-VM                     & 0.176                   & 0.106                   & 0.173                   & 0.338                   & 0.187                   & 0.076                   \\
NeRF                     & 0.198                   & 0.111                   & 0.192                   & \textbf{0.395}          & 0.196                   & 0.098                   \\
s-MLP                    & \textbf{0.194}          & \textbf{0.102}          & \textbf{0.182}          & 0.418                   & \textbf{0.190}          & \textbf{0.081}          \\
Plenoxels128             & \textbf{0.219}          & \textbf{0.128}          & \textbf{0.231}          & \textbf{0.394}          & \textbf{0.240}          & \textbf{0.106}          \\
s-Tensors                & 0.263                   & 0.158                   & 0.268                   & 0.488                   & 0.289                   & 0.114                   \\ \hline
                         & \multicolumn{6}{c}{$\text{LPIPS}_{Vgg}$}                                                                                                                             \\ \hline
\textit{\textbf{t-Hash}} & \textit{\textbf{0.134}} & \textit{\textbf{0.081}} & \textit{\textbf{0.132}} & \textit{\textbf{0.244}} & \textit{\textbf{0.161}} & \textit{\textbf{0.056}} \\
TensoRF-VM               & \textbf{0.155}          & \textbf{0.085}          & \textbf{0.161}          & \textbf{0.278}          & \textbf{0.177}          & 0.074                   \\
s-VM                     & 0.181                   & 0.125                   & \textbf{0.161}          & 0.375                   & 0.178                   & \textbf{0.067}          \\
NeRF                     & -                       & -                       & -                       & -                       & -                       & -                       \\
s-MLP                    & 0.190                   & 0.094                   & 0.195                   & 0.371                   & 0.200                   & 0.091                   \\
Plenoxels128             & 0.261                   & 0.160                   & 0.263                   & 0.474                   & 0.287                   & 0.125                   \\
s-Tensors                & \textbf{0.220}          & \textbf{0.125}          & \textbf{0.234}          & \textbf{0.398}          & \textbf{0.242}          & \textbf{0.101}          \\ \hline
\end{tabular}
}
\caption{Comparing the per-scene metrics of models (s-VM, s-MLP, s-Tensors) obtained by PVD with the models (TensoRF-VM, NeRF, Plenoxels) trained from scratch on TanksAndTemples dataset.}
\label{tab-gp2others-tanks}
\end{table*}

% clipVSnoclip - loss density and RGB

% clipVSnoclip - loss density and RGB

% clipVSnoclip - loss density and RGB

% clipVSnoclip - loss density and RGB

% clipVSnoclip - loss density and RGB

% clipVSnoclip - loss density and RGB

\section{Acknowledgement}
This work is supported by the National Natural Science Foundation of China (U20B2042, 62076019), Science and Technology Innovation 2030-Key Project of “New Generation Artificial Intelligence”(2020AAA0108201).

%\label{table-bloackwise-ablation}
%\caption{An ablation study of the block-wise strategy. Metrics are averaged over the 8 scenes from  NeRF-synthetic dataset. vm to mlp}

% Using the \centering command instead of \begin{center} ... \end{center} will save space
% Positioning your figure at the top of the page will save space and make the paper more readable
% Using 0.95\columnwidth in conjunction with the

% Use \bibliography{yourbibfile} instead or the References section will not appear in your paper
%\nobibliography{aaai23}
\bibliography{aaai23}
%\bibliography{aaai23.bib}

\end{document}